\pdfoutput=1

\documentclass[11pt]{article}

\usepackage[preprint]{acl}

\usepackage{times}
\usepackage{latexsym}
\usepackage{booktabs}
\usepackage{multirow}
\usepackage{adjustbox}
\usepackage{todonotes}
\usepackage[T1]{fontenc}
\usepackage[utf8]{inputenc}
\usepackage{microtype}
\usepackage{inconsolata}
\usepackage{microtype}
\usepackage{inconsolata}
\usepackage[most]{tcolorbox}
\usepackage[dvipsnames]{xcolor}
\usepackage{arydshln}
\usepackage{wrapfig}
\usepackage{enumitem}

\usepackage{graphicx}

\newtcolorbox{prompt}[3][]{%
  enhanced,
  breakable,
colback=blue!5!white,
colframe=blue!75!black,
  title=\textbf{ \thetcbcounter#3},
  fontupper=\footnotesize\fontfamily{ttfamily}\selectfont,
  #1
}
\usepackage{tikz}
\usepackage{forest}
\usetikzlibrary{trees,positioning,shapes,shadows,arrows.meta}
\usepackage{multicol}

\title{\textsc{AfriLangTutor}: Advancing Language Tutoring and Culture Education in Low-Resource Languages with Large Language Models}

\author{\normalsize Tadesse Destaw Belay$^{1}$, Shahriar Kabir Nahin$^{2}$, Israel Abebe Azime$^{3}$, Ocean Monjur$^{2}$\\ 
\textbf{\normalsize Marek Rei$^{4}$, Chris Biemann$^{5}$, Shamsuddeen Hassan Muhammad$^{4}$, } \\
\textbf{\normalsize Seid Muhie Yimam$^{5}$, Anshuman Chhabra$^{2}$ } \\
\footnotesize
 $^1$Instituto Politécnico Nacional, Mexico,  $^2$University of South Florida, FL, USA, $^3$Saarland University, Germany,
 \\ 
 \footnotesize
$^4$Imperial College London, UK,  $^5$University of Hamburg, Germany
\\}

\begin{document}
\maketitle

\begin{abstract}

\looseness-1 \textit{How can language learning systems be developed for languages that lack sufficient training resources?} This challenge is increasingly faced by developers across the African continent who aim to build AI systems capable of understanding and responding in local languages. To address this gap, we introduce \textsc{AfriLangDict}, a collection of \textbf{194.7K} African language-English dictionary entries designed as seed resources for generating language-learning materials, enabling us to automatically construct large-scale, diverse, and verifiable student–tutor question–answer interactions suitable for training AI-assisted language tutors. Using \textsc{AfriLangDict}, we build \textsc{AfriLangEdu}, a dataset of \textbf{78.9K} multi-turn training examples for Supervised Fine-Tuning (SFT) and Direct Preference Optimization (DPO). Using \textsc{AfriLangEdu}, we train language tutoring models collectively referred to as \textsc{AfriLangTutor}. We fine-tune two multilingual LLMs: Llama-3-8B-IT and Gemma-3-12B-IT on \textsc{AfriLangEdu} across 10 African languages and evaluate their performance. Our results show that models trained on \textsc{AfriLangEdu} consistently outperform their base counterparts, and combining SFT and DPO yields substantial improvements, with gains ranging from 1.8\% to 15.5\% under LLM-as-a-judge evaluations across four criteria. To facilitate further research on low-resource languages, all resources are available at \url{https://huggingface.co/afrilang-edu}. 

\end{abstract}

\section{Introduction}

\looseness-1 Large Language Models (LLMs) have demonstrated significant progress in downstream natural language processing (NLP) tasks, enabling human-like language understanding and data generation across diverse domains, such as education \cite{chu-etal-2025-llm}, medical applications \cite{maity2025large}, and others \cite{10.1145/3735632}.
However, their performance is heavily constrained by the volume of the specific language data on which they are pre-trained \cite{muennighoff2025scalingdataconstrainedlanguagemodels}. For low-resource languages (LRLs), limited training coverage leads to weak lexical knowledge and unreliable linguistic interpretations when models are asked to generate or explain words in those local languages \cite{pucinskaite-mitkov-2025-evaluating}. On the other hand, for high-resource languages (e.g., English), owing to the large volumes of pre-training data, LLMs demonstrate strong linguistic understanding and can effectively serve as language tutors in these languages \cite{ye-etal-2025-position}. However, their performance as tutors in LRLs remains largely unexplored. 

\begin{figure}[t]
    \centering
    \includegraphics[width=1\linewidth]{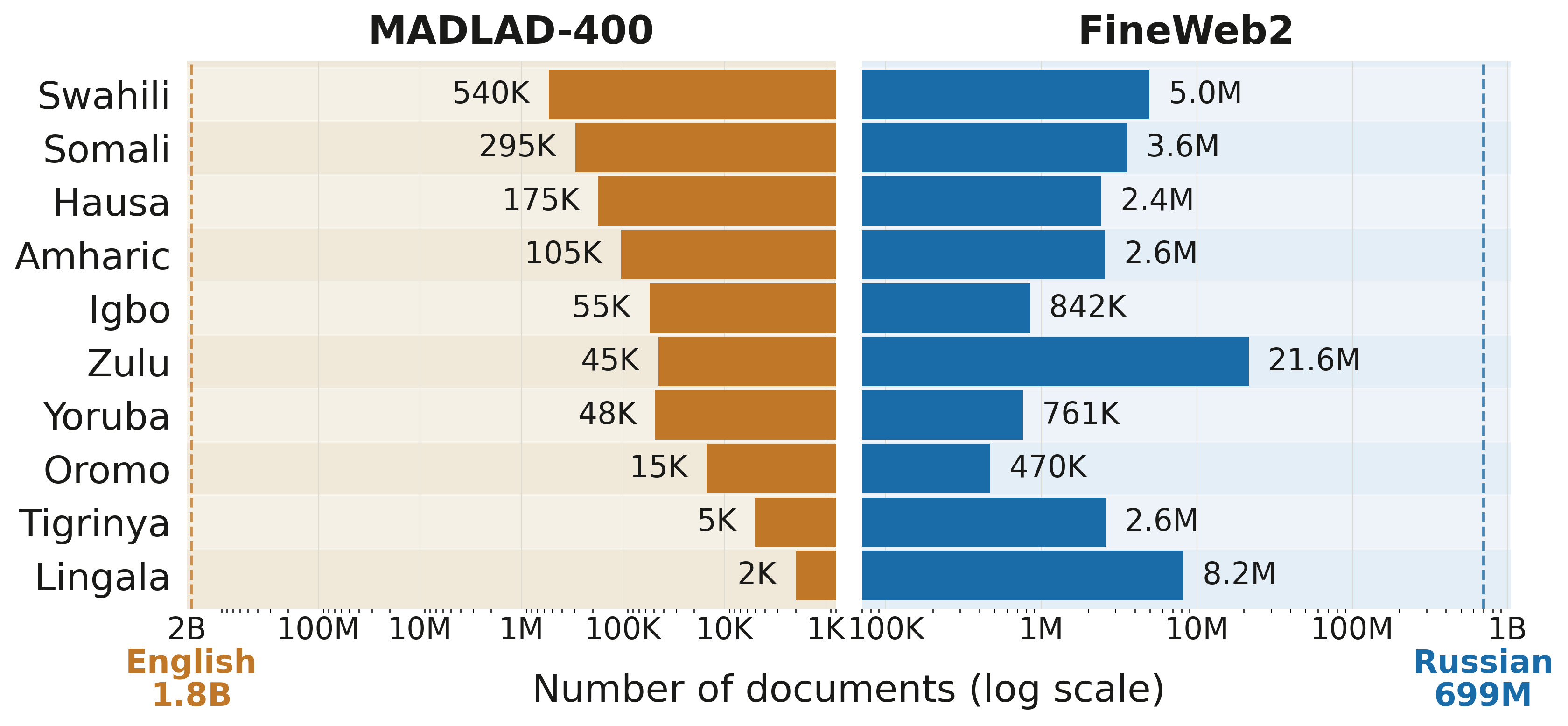}\vspace{-3mm}
    \caption{\textbf{Number of documents for 10 African LRLs} in two widely used pretraining corpora: MADLAD-400 (left) and FineWeb2 (right), compared with high-resource: English (1.8B) and Russian (699M).}
    \label{fig:LRL_data}\vspace{-6.5mm}
\end{figure}

\looseness-1 Figure~\ref{fig:LRL_data} compares the number of documents for 10 African LRLs with high-resource reference languages in two widely used pretraining corpora: MADLAD-400~\cite{NEURIPS2023_d49042a5} and FineWeb2~\cite{penedo2025fineweb2}. African LRLs contain far fewer documents than high-resource languages such as English (1.8B in MADLAD-400) and Russian (699M in FineWeb2), and in some cases account for less than 0.01\% of the English baseline. This severe imbalance highlights a fundamental limitation of multilingual pretraining corpora and underscores the difficulty of training models that generalize effectively to African LRLs.

\looseness-1 Synthetic data generation is often proposed as a solution for data scarcity in many languages \cite{de-gibert-etal-2025-scaling,anikina-etal-2025-rigorous}. However, the quality of the generated content can be highly dependent on the context (prompt) or seed data used during generation \cite{yong-etal-2024-lexc}. Structured seed resources can act as stable anchors that guide generation and improve the utility of synthetic examples. Among linguistic resources, dictionaries and bilingual lexicons are both \textit{foundational} and relatively \textit{accessible}, even for extremely low‑resource languages, making them especially suitable as seeds for data generation \cite{alam2024morphologicallyaware}. By using dictionary entries as a foundation for LLMs' synthetic data generation, it becomes possible to expand from individual words to phrases, sentences, and more complex linguistic structures in a controlled and reliable manner \cite{long-etal-2024-llms}.

In this work, we prepare new bilingual parallel dictionaries (LRLs--English) for 10 African languages called \textsc{AfriLangDict}. We then use these dictionaries to generate pedagogically useful language tutoring materials called \textsc{AfriLangEdu}, using various question-and-answer templates. This \textit{word-focused} and \textit{context-aware} generation reduces hallucinations and produces more accurate, culturally grounded language-tutoring content. 
  
Our work thus seeks to answer the following Research Questions (RQs): \textbf{(1)} To what extent can state-of-the-art LLMs effectively function as tutors for low-resource languages (LRLs)? \textbf{(2)} Can dictionary-based seed resources be leveraged to generate pedagogically meaningful, multi-turn tutoring dialogues across typologically diverse languages? \textbf{(3)} How effectively can dictionary-driven multilingual tutoring LLMs support scalable language learning for LRLs? 
\textbf{(4)} What is the effectiveness of alignment techniques, such as Supervised Fine-Tuning (SFT) \cite{NEURIPS2022_b1efde53}, Direct Preference Optimization (DPO) \cite{rafailov2023direct}, and their combination (SFT + DPO), in specializing LLMs for multilingual language tutoring?

\noindent In sum, our main contributions are as follows:
\begin{itemize}
    \item We introduce: (1) \textsc{AfriLangDict}, constituting a new dictionary of entries (194K) for 10 African languages; (2) \textsc{AfriLangEdu}, comprising a new synthetic dataset (78.9K multi-turn and DPO data) that is generated using \textsc{AfrilangDict} as seed data from the state-of-the-art LLMs; and (3) \textsc{AfriLangTutor}, open-source language tutoring LLMs trained and localized using \textsc{AfriLangEdu}.
    \item We evaluate existing LLMs for their effectiveness as language tutors, assess alignment techniques (e.g., SFT, DPO, and SFT + DPO) across multiple evaluation metrics (e.g., automatic metrics, LLM-as-a-judge, and human evaluation), and identify significant opportunities to improve African language education. 
\end{itemize}

\begin{table*}[t!]
\centering

\resizebox{\textwidth}{!}{
\begin{tabular}{cl ll rrrr}

\toprule
\# & \textbf{Language} (\textbf{ISO}) & \textbf{Script} & \textbf{Spoken in} & \textbf{AfriLangDict}& \textbf{Multi-turn SFT} & \textbf{DPO} & \textbf{Test} \\
\midrule
1 & Amharic (\texttt{amh})  &Ethiopic& Ethiopia, Eritrea  & 13,621 & 4,599 & 3,002&1,000\\
2 & Hausa (\texttt{hau}) & Arabic/Latin& Northern Nigeria, Ghana, Cameroon  & 7,449& 3,661  & 3,054 &1,002 \\
3 & Igbo (\texttt{ibo}) &Latin  & Southeastern Nigeria    &18,992& 3,697 & 3,584&1,032\\
4 & Lingala  (\texttt{lin}) &Latin&Congo, Central African    &6,712&3,600 &3,000&1,000\\
5 & Oromo (\texttt{orm})  &Latin&Ethiopia, Somalia   & 15,073&3,600   &3,000&1,000\\
6 & Somali (\texttt{som}) &Latin& Ethiopia, Kenya, Somalia   & 12,577& 3,600 &3,000&999\\
7 & Swahili (\texttt{swa}) &Latin&Tanzania, Kenya, Mozambique   &27,468 & 3,600&2,871&1,052\\
8 & Tigrinya (\texttt{tir}) &Ethiopic&Ethiopia, Eritrea  & 30,704& 3,625&31.98&1,000\\
9 & Yoruba (\texttt{yor}) &Latin&Southwestern, Central Nigeria, Togo     &28,591& 4,113 &3,325&999\\
10 & Zulu (\texttt{zul}) &Latin &South Africa    &33,605&3,600   &3,000&1,059\\
\hline
 &  &  & & \textbf{Total} = \textbf{194,792}&\textbf{37,695}&\textbf{31,034} &\textbf{10,143}\\
\bottomrule
\end{tabular}
}
\caption{\textbf{Overview of the 10 African languages in \textsc{AfriLangDict} and \textsc{AfriLangEdu}}, including script, region, dictionary size, and the number of generated multi-turn SFT, DPO, and test instances per language.}
\label{tab:dict}

    
\end{table*} 

\section{Related Work}

\subsection{Large Language Models for Education}
\looseness-1 In recent years, LLMs have been increasingly applied in many fields, including recommendation, government, education, legal affairs, and
finance \cite{xu2024largelanguagemodelseducation}. LLM has advanced education in many aspects, including personalized learning support \cite{liu2025beyond}, interdisciplinary capabilities \cite{LIU2025100741}, real-time problem-solving and tutoring \cite{dinucu-jianu-etal-2025-problem}, and broader educational knowledge coverage \cite{ahmad-genai-2023}. Although LLMs have shown strong potential to improve teaching, change educational models, and support teachers, they still face major challenges, especially in LRL settings. These include lower performance and higher error rates in LRLs compared to high-resource languages \cite{StanfordHAI2025}, cultural misalignment due to Western-centric training data \cite{Tao2024}, and the widening digital divide caused by limited infrastructure and computational access in the Global South \cite{Mokoena2025}. 

\subsection{Language Tutoring in the Era of LLMs}

\looseness-1 \paragraph{Dictionary-centric Approaches.} 
Language dictionaries are a foundational resource for building language models and enabling multilingual understanding \cite{sakajo-etal-2025-dictionaries}. 
Bilingual dictionaries support a wide range of tasks, including enhancing rare words translation quality \cite{goyal2025iolbench}, alignment methods \cite{gaschi-etal-2023-exploring}, and cross-lingual transfer \cite{sakajo-etal-2025}. Additionally, \citet{zhang-etal-2024-teaching} demonstrates that LLMs can be effectively adapted to unseen words in LRLs through in-context learning using a parallel dictionary for machine translation.

Beyond direct translation, dictionaries have been integrated into more sophisticated LLM systems to guide and constrain model behavior. Recent dictionary-augmented frameworks refine LLM queries before execution by injecting lexical and semantic priors from bilingual lexicons. These priors support several functions, including handling rare and unseen words during prompting \cite{lu-etal-2024-chain,yin-etal-2024-lexmatcher}, curating higher-quality machine translation data \cite{yin-etal-2024-lexmatcher}, and enabling dictionary-aware prompting strategies that improve alignment between user intent and model outputs \cite{CAO2023110605}. \citet{goyal2025iolbench} introduce the International Linguistics Olympiad (IOL) benchmark by constructing parallel dictionaries and formulating translation problems in which random words or phrases from either side are presented as self-contained linguistic puzzles to evaluate the reasoning abilities of LLMs.

Overall, dictionary-based approaches offer several advantages for LLM-driven applications, particularly in low-resource settings. They help reduce hallucinations by grounding model outputs in verified lexical knowledge, filter irrelevant or noisy queries, and enable continuous improvement through user feedback and query history \cite{gao2024retrieval}. These properties make dictionaries a powerful and practical tool for extending LLMs to LRLs and supporting culturally grounded language use.

\paragraph{LLMs for Language Proficiency Assessment.}
Recent work has demonstrated the growing role of LLMs in evaluating and supporting language learning. \citet{xu-etal-2025-large} introduces the Chinese Language Teaching Evaluation (CLTE) benchmark, which assesses linguistic competence, cultural knowledge, and instructional quality in Chinese language education. Similarly, \citet{imperial-etal-2025-universalcefr} propose the Multilingual UNIVERSALCEFR benchmark, which provides Common European Framework of Reference for Languages (CEFR) level annotations from A1 to C2 for standardized proficiency classification.

\looseness-1 Beyond proficiency labeling, several benchmarks focus on LLMs as active language tutors. \citet{srinivasa2025tutorbenchbench} introduces TUTORBENCH, a comprehensive evaluation framework that measures core tutoring capabilities of LLMs, including generating adaptive explanations based on student misunderstandings, providing actionable feedback on learner outputs, and promoting active learning through effective hint generation. In addition, large-scale analyses of conversational agents for language learning show that chatbots can positively impact second language acquisition \cite{Effectiveness2025}. Some position papers and empirical studies further argue that LLMs can serve as effective tutors in English education by complementing human instructors and mitigating limitations of traditional classroom settings \cite{ye-etal-2025-position,karatacs2024incorporating}.

Despite these advances, the potential of LLMs as language tutors remains largely unexplored in pedagogically challenging, morphologically complex settings, especially for low-resource African languages learned as second languages. Our work addresses this gap by advancing multilingual AI-assisted language education systems that aim to approach the effectiveness of human instructors while supporting culturally and linguistically diverse LRL learners.

\section{Dataset and Construction Details}
\begin{figure*}[!h]
    \centering
    \includegraphics[width=0.98\linewidth]{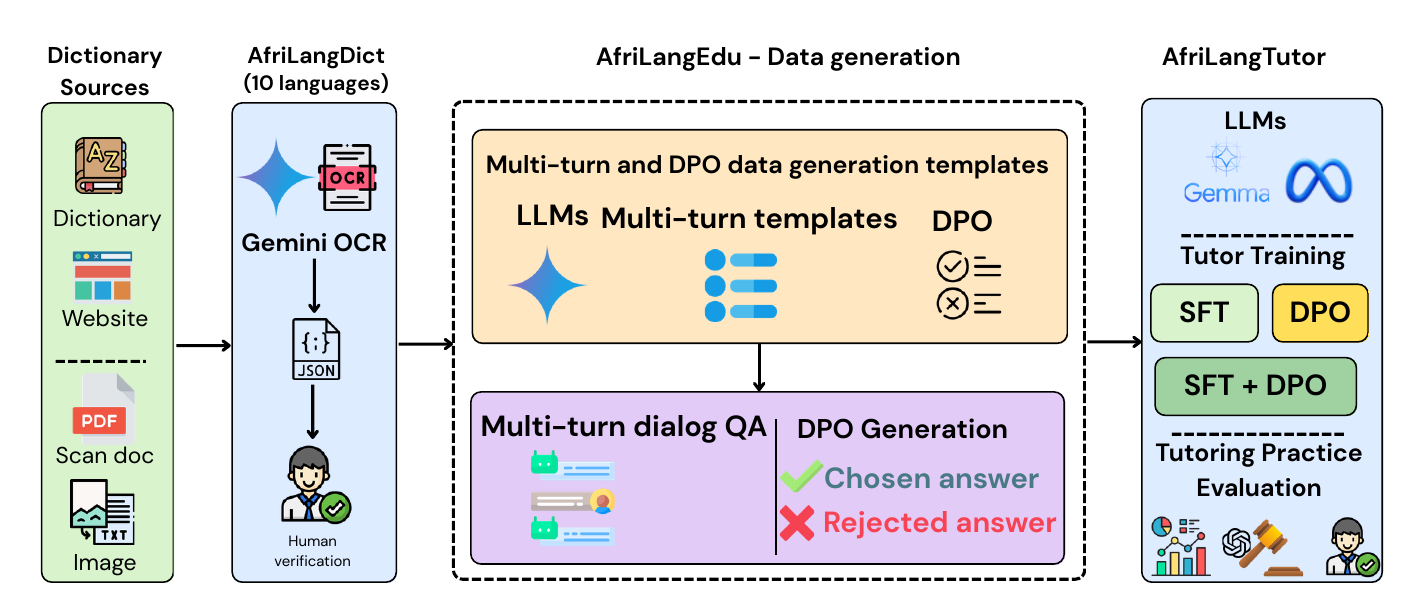}\vspace{-2mm}
    \caption{\textbf{Overview of the \textsc{AfriLangTutor} pipeline.} Dictionary sources are collected and processed via OCR and human verification to construct \textsc{AfriLangDict} across 10 languages. These entries serve as seed data for synthetic generation of \textsc{AfriLangEdu}, which comprises multi-turn tutoring dialogues and DPO preference pairs. Finally, Llama-3-8B and Gemma-3-12B are fine-tuned using SFT, DPO, and SFT+DPO to produce the \textsc{AfriLangTutor} models.}
     \label{fig:pipline}\vspace{-3mm}

\end{figure*}

\subsection{\textsc{AfriLangDict}: Dictionary Collection}
\looseness-1 Among the world’s 7,000 languages, 95\% lack sufficient data (>100K sentences) to train LLMs \cite{bapna2022buildingmt}. However, most languages have a grammar book (60\%) or a dictionary (75\%) \cite{nordhoff2011glottolog}, including many endangered low-resource languages. Dictionaries are relatively easy to obtain, even for LRLs, making them appealing candidates for many downstream NLP tasks.  Therefore, we begin by collecting an African-language dictionary (\textsc{AfriLangDict}) and use these linguistic resources to enable LLMs to generate educational resources and assist with LRL tutoring.\vspace{1.5mm}

\looseness-1\noindent\textbf{Dictionary Sources.}
To create \textsc{AfriLangDict}, we draw on several sources of bilingual dictionaries. These include scanned PDF dictionaries, which we convert into a standardized machine-readable format using OCR tools such as the Google Cloud Vision API\footnote{\scriptsize{\url{https://cloud.google.com/vision/docs/ocr}.}} and PyPDF2\footnote{\scriptsize{\url{https://pypi.org/project/PyPDF2/}.}}, as well as online dictionary platforms from which we scrape bilingual entries, specifically Abyssinica\footnote{\scriptsize{\url{https://dictionary.abyssinica.com/}.}} for Amharic and IgboGuide\footnote{\scriptsize{\url{https://www.igboguide.org/HT-vocabulary.htm}.}} for Igbo. All extracted entries are normalized into a unified JSON format and subsequently verified by native speakers to ensure accuracy.

\noindent\textbf{Language Selection.} 
We target 10 African languages based on the availability of dictionary resources. These languages include various scripts (Latin, Arabic, Ge'ez (Ethiopic)) and language families (Afro-Asiatic: Amharic, Oromo, Hausa, Somali, Tigrinya, and Niger-Congo: Igbo, Lingala, Swahili, Yoruba, Zulu). We preprocess the dictionary entries to ensure proper Markdown formatting, preserving compatibility with special symbols, underlines, and other formatting conventions commonly found in educational materials. Finally, the dictionary entry is verified by a native speaker for each language to check correctness, alignment with English meaning, and correction of OCR errors. The train-test split began at the AfriLangDict level. Details of the languages considered, along with the statistics of \textsc{AfriLangDict} and \textsc{AfriLangEdu} data (multi-turn for SFT fine-tuning, the DPO data for alignment training, and test set) are presented in Table \ref{tab:dict}. Figure \ref{fig:pipline} illustrates our general data creation pipeline.

\subsection{\textsc{AfriLangEdu}: Data Generation}
Using \textsc{AfriLangDict}, we generate multi-turn dialogues for SFT and DPO data to align subjective preferences (for helpfulness, teaching style, and response tone optimization). %

\begin{figure*}[h!]
    \centering
    \includegraphics[width=0.91\linewidth]{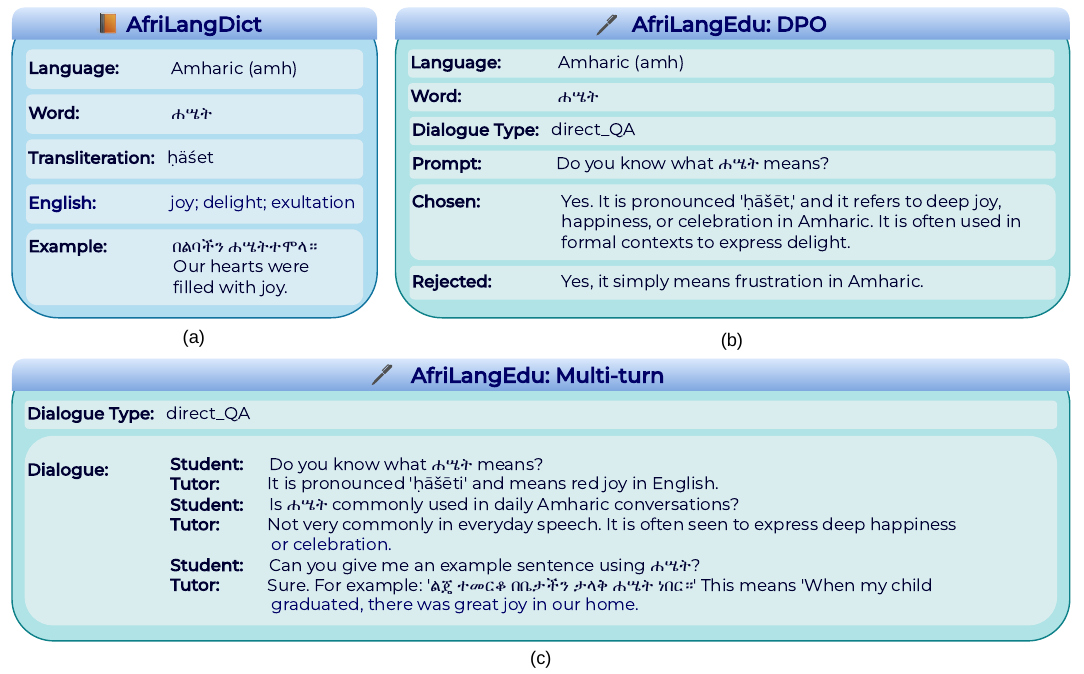}\vspace{-2.5mm}
    \caption{\textbf{Data format and examples:} (a) \textsc{AfriLangDict} dictionary format, (b) DPO data, and (c) multi-turn dialog with 3 full turns. Both (b) and (c) comprise \textsc{AfriLangEdu} and are generated using \textsc{AfriLangDict}. The multi-turn responses and the chosen answer for DPO is generated using the highly performant Gemini-2.5-Pro \cite{comanici2025gemini}, and the rejected answer for DPO is generated using various lower LRL quality open-source LLMs of differing sizes (e.g. Llama-3 (1B, 8B) \cite{grattafiori2024llama3herdmodels} and Gemma-2-2B \cite{gemmateam2024gemma2improvingopen}).}
    \label{fig:data-format}\vspace{-2mm}
\end{figure*}
\subsubsection{Multi-turn Dialog Generation}
A limitation of many existing open-source datasets is that they are unstructured or consist of single-turn question-answer pairs, and they often lack natural seed data when synthetically generated, which limits diversity and can introduce bias \cite{Towardsbias2025}. Unlike single-turn question-answer data, multi-turn data captures the dynamic nature of human conversations, enabling models to learn to maintain context, follow conversational flow, adapt to changes in user intent, and engage in complex interactions \cite{chen2025learning,li2026singleturnsurvey}. We generate language and culture-centered multi-turn tutoring data using \textsc{AfriLangDict} as seed data with Gemini-2.5-Pro \cite{comanici2025gemini}. We design multiple generation templates covering tasks such as direct question answering, contextual dictionary use, sentence construction, translation practice, cultural note integration, spelling and pronunciation, and more. We used this dictionary-based, generated multi-turn data for the SFT training, described in subsequent sections.  

\subsubsection{DPO Data Generation} 
\looseness-1 Preference optimization via methods such as DPO \cite{rafailov2023direct}, has been widely adopted as a standard technique for aligning LLMs with human preferences \cite{pant2025improvingllmsafetyhelpfulness}. In addition to multi-turn data, we generate DPO data with \textbf{chosen} (preferred) and \textbf{rejected} (less preferred) responses using specific pre-defined templates (details are provided in Appendix \ref{app:temlates}). More specifically, the DPO training data consists of paired input–output examples that capture both correct and incorrect interactions. These pairs are constructed to reflect different combinations of query and response quality: (a) \textbf{\textit{Correct input and response}}: both the learner’s query and the tutor’s response are appropriate and beneficial for the language learner. (b) \textbf{\textit{Incorrect query with correct response}}: the learner’s query contains linguistic or conceptual issues, but the tutor’s response provides a correct explanation or correction. 
(c) \textbf{\textit{Correct query with incorrect response}}: the learner’s query is valid, but the generated response is incorrect, misleading, or outside the intended scope. The multi-turn SFT data and the chosen responses for DPO (correct responses to both correct and incorrect queries) are generated using Gemini-2.5-Pro. The rejected responses (incorrect responses to both correct and incorrect queries) are generated using small versions of Llama-3 and Gemma-2 LLMs. An example of the data formats and structure of \textsc{AfriLangDict} and \textsc{AfriLangEdu} are shown in Figure \ref{fig:data-format}.

\subsection{Data Quality Control}
\begin{figure*}[h]
    \centering
    \includegraphics[width=1\linewidth]{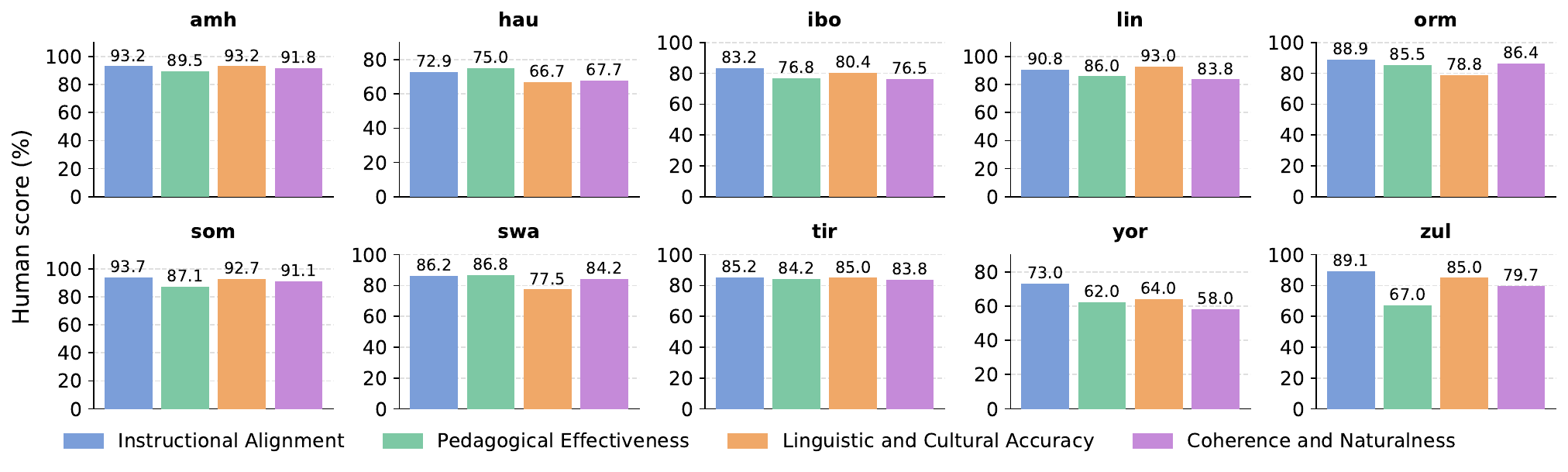}\vspace{-3mm}
    \caption{ Human validation score (\%): across four judgment criteria (details in Sec. \ref{sec:eval}), from $N=100$ samples. }
    \label{fig:hum-eval}\vspace{-2.5mm}
\end{figure*}

\paragraph{Human Validation} To ensure data quality, we have had native speakers validate random samples for each language using four judgment criteria that we will also use later for LLM-as-a-judge evaluation. Following prior work that uses human verification for LLM-generated outputs, we selected 100 random samples (10\% of a language test set) as a practical compromise between verification burden and confidence in data quality \cite{chen2024llmgenerated}. As can be observed, the human validation score in Figure \ref{fig:hum-eval} indicates acceptable quality of the generated text.

\paragraph{Influence Analysis}
We analyze influential samples from the SFT training data based on their influence scores \cite{pruthi2020estimatingtrainingdatainfluence}. Influence analysis quantifies how individual training samples affect model predictions on a given validation set, allowing us to identify both beneficial and detrimental examples \cite{ICLR2024_04cda3a5, chhabra2025oga}. By ranking samples according to their influence, where a positive score indicates a beneficial sample, we can prioritize high-impact data for model fine-tuning and reduce the effect of harmful examples, improving overall alignment and generalization. We performed influence analysis on our SFT dataset using the fine-tuned Llama-3-8B-IT model. The results in Section~\ref{sec:results} show that all samples have positive influence scores, indicating that each sample contributes positively to the model’s performance. Further details of influence analysis are provided in Appendix \ref{app:influence-analysis}.

\section{Experimental Set-up}

\subsection{\textsc{AfriLangTutor}: Training LLMs}
\looseness-1 We fine-tune two open-source LLMs, Llama-3-8B and Gemma-3-12B, using SFT and DPO to develop \textsc{AfriLangTutor}. SFT trains the models on our multi-turn tutoring data. DPO further refines the models using pairwise preference data (i.e., \textit{chosen} and \textit{rejected} responses), encouraging the model to generate preferred outputs. Note that we apply SFT and DPO independently of each other, and also in combination (i.e., SFT + DPO) to better understand how each training paradigm affects downstream task performance. 
We fine-tune the LLMs with several varying hyperparameters and provide additional implementation details in Appendix \ref{app:param}. 

\begin{table*}[!h]
\centering

\resizebox{0.88\textwidth}{!}{
\begin{tabular}{lccccccccccc}
\toprule
\textbf{Model} & \textbf{amh} &\textbf{hau}& \textbf{ibo}& \textbf{lin}& \textbf{orm}& \textbf{som}& \textbf{swa}& \textbf{tir}& \textbf{yor}& \textbf{zul}& \textbf{Avg}\\
\hline
\multicolumn{10}{l}{\textit{African Language-Centric LLMs}} \\
AfriqueGemma-4B & 33.3&	33.6&	32.0&	31.2&	35.9&	32.1&	32.7&	33.2&	31.7&	33.1&	32.6\\
AfriqueLlama-8B & 35.7&	34.8&	32.0&	31.2&	45.3&	33.1&	34.1&	35.7&	32.0&	32.5&	33.4\\
AfriqueGemma-12B & 31.4&	33.0&	33.3&	30.8&	31.7&	31.9&	33.2&	32.2&	32.5&	32.4&	32.3\\
AfriqueQwen-14B & 41.5&	40.2&	43.6&	39.2&	40.0&	40.6&	38.8&	41.5&	40.6&	38.7&	40.5\\
\hline
Gemma-2-2B & 51.3&	59.3&	56.2&	58.3&	58.7&	56.2&	59.6&	52.5&	57.9&	60.5&	57.1 \\
Llama-3.2-1B & 44.7&	52.6&	51.48&	51.7&	53.8&	51.0&	50.4&	46.2&	51.5&	53.6&	50.7 \\
Llama-3.2-3B & 47.5&	60.0&	56.6&	58.2&	58.4&	55.0&	58.7&	49.5&	56.1&	60.7&	56.1\\
Aya-global-3.5B & 55.9&	57.2&	60.0&	56.2&	53.8&	53.5&	59.6&	55.2&	57.3&	56.9&	56.6 \\
Gemma-3-4B & 53.8&	55.3&	54.3&	56.6&	54.5&	54.5&	57.1&	54.2&	54.3&	55.1&	55.0 \\
Qwen3-4B & 48.1&	57.1&	55.0&	55.9&	57.4&	54.7&	54.4&	48.5&	55.1&	57.4&	54.4\\
Llama-3-8B & 53.0&	60.2&	58.6&	60.7&	59.5&	57.1&	61.0&	54.2&	59.5&	60.6&	\textbf{58.5}\\
Gemma-3-12B & 59.9&	60.8&	59.1&	60.0&	58.2&	60.6&	63.2&	59.4&	58.5&	60.0&	\textbf{59.9} \\

\bottomrule
\end{tabular}
}
\caption{Zero-shot test-set benchmarking of LLMs for LRLs tutoring. The results are percentage results averaged across the four judging criteria as described in Section \ref{sec:eval}. Note that Llama-3-8B and Gemma-3-12B attain the highest zero-shot performances, and are thus used to obtain our \textsc{AfriLangTutor} models subsequently.}
\label{tab:benchmark}\vspace{-2mm}

\end{table*}

\subsection{Evaluation Metrics}\label{sec:eval}
\looseness-1\noindent \textbf{Automated Evaluation Metrics}
We use general domain-agnostic reference overlap metrics such as ChrF++ \cite{popovic-2017-chrf},  BERTScore \cite{Zhang2020BERTScore}, and ROUGE-L \cite{lin-2004-rouge} as proxies to assess the coherence and human-likeness of AI tutor responses \cite{liu-etal-2023-g}. \vspace{1.5mm}

\looseness-1\noindent\textbf{LLM-as-a-Judge} 
In practice, language tutoring requires more than exact word matching, as commonly measured by overlap metrics (e.g. BERTScore F1, ChrF++, and ROUGE-L). A tutor may explain a concept using a valid alternative phrasing that does not match the reference answer and automated metrics, hence, unable to capture pedagogical quality or instructional usefulness. As evaluation of generated text is increasingly shifting toward LLM-as-a-judge frameworks rather than solely relying on automatic metrics \cite{kochmar-etal-2025-findings}, we additionally report LLM-as-a-judge evaluation results. We select \textbf{GPT-5.2} for LLM-as-a-judge because of GPT family models perform better than other LLMs in the AfroBench leaderboard \cite{ojo-etal-2025-afrobench}. We adopt the three LLM-as-a-judge evaluation criteria from \citet{singh2026tinyaya} and introduce a new criterion, "Pedagogical Completeness," tailored to language tutoring evaluation. Additional details are provided in Appendix \ref{app:llm-judge-prompt}. We now provide the evaluation criteria for the LLM judge:\vspace{-3.5mm}
\begin{itemize}[noitemsep]
    \item \textit{Tutoring Instruction Alignment}: Ensure that the model correctly identifies the user's specific learning intent and adheres to all task-related constraints like format, target language, and difficulty level. It prevents the tutor from providing irrelevant information or ignoring the learner's requested activity type.
    \item \textit{Pedagogical Completeness}:  Evaluate if the response provides a comprehensive learning package, including clear explanations, contextual examples, and supportive scaffolding. It distinguishes between a simple "correct answer" and a high-quality educational resource that guides a student toward mastery.
    \item \textit{Linguistic and Cultural Accuracy}: Validate whether the language is grammatically flawless and reflects cultural authenticity. It ensures learners are taught nuances, such as social honorifics and idiomatic expressions, that are essential for real-world communication.
    \item \textit{Coherence and Naturalness}: Assess the logical flow and the professional tutoring persona, ensuring the text is organized, encouraging the learner, and easy to follow. 
\end{itemize}\vspace{-3.5mm}

\section{LLMs for Language Tutoring}
\subsection{Baselines} 
\looseness-1 To assess how state-of-the-art LLMs perform as tutors for LRLs, we benchmarked the latest multilingual and open-source LLMs, including Gemma-2-2b-IT \cite{gemmateam2024gemma2improvingopen}, Gemma-3 (4B, 12B) \cite{kamath2025gemma}, Qwen3-4B \cite{yang2025qwen3}, Llama-3 (1B, 3B, 8B) \cite{grattafiori2024llama3herdmodels}, and Tiny-Aya-global-3.5B \cite{singh2026tinyaya}. In addition, we evaluate the recent African-centric LLMs such as AfriqueLLM (Gemma-(4B,12B), Llama-8B, and Qwen-14B) \cite{yu2026afriquellmdatamixingmodel}. All models are instruction fine-tuned versions for fair comparison. We selected these LLMs to facilitate efficient computation, support multilingual capabilities, and ensure reproducibility. 

\subsection{Benchmarking, Results, and Analysis}\label{sec:results}
We first conduct zero-shot prompting as a benchmark and then in subsequent sections, aim to fine-tune the best LLMs with SFT, DPO, and SFT + DPO to create our \textsc{AfriLangTutor} models. Using GPT-5.2 as the LLM-as-a-judge, we present zero-shot performance in Table \ref{tab:benchmark}. As can be observed, the African-centric LMs perform worse than the base models, and even against smaller-sized LLMs such as Llama-3.2-1B and Gemma-2-2B. We posit this is due to: 1) the domain of the training data, which includes code, mathematics, and synthetic data \cite{yu2026afriquellmdatamixingmodel}, and 2) while the AfriqueLLM models are aligned specifically with African languages during continual pre-training, our task focuses on English-centric-targeted English speaker language learning, leading to a domain mismatch. Note that the larger-sized LLMs, such as Llama-3-8B and Gemma-3-12B, tend to perform better than other lower parameter count models. Thus, we will use these in the next section for further fine-tuning. Interestingly, among the smaller LLMs, Gemma-2-2B outperforms higher-parameter models such as Llama-3.2-3B, Qwen3-4B, and Tiny-Aya-global-3.5B. We posit this might be due to more extensive training for Gemma-2-2B on multilingual data. 

\subsection{Improving LLMs for Language Tutoring}
To improve LLMs for tutoring LRLs, we evaluate the effectiveness of training via SFT, DPO, and their combination. Here we present ChrF++  and LLM-as-a-judge (GPT-5.2) results in Table~\ref{tab:result1}. Results for BERTScore and ROUGE-L as the evaluation metrics are deferred to Appendix \ref{app:auto-metrics} due to space constraints. We provide additional ablations with different fine-tuning parameters, DPO rejected response variants, and the impact of full weight and LoRA-based fine-tuning results in Appendix \ref{app:addional-results}. 
From the results obtained in Tables~\ref{tab:result1}, we can ascertain the following:

\begin{table*}[h!]
\centering

\resizebox{0.9\textwidth}{!}{
\begin{tabular}{crcccccccccc|c}
\toprule
\textbf{Metric}&\textbf{Model} & \textbf{amh} &\textbf{hau}& \textbf{ibo}& \textbf{lin}& \textbf{orm}& \textbf{som}& \textbf{swa}& \textbf{tir}& \textbf{yor}& \textbf{zul}& \textbf{Avg}\\
\hline
\multirow{8}{*}{\rotatebox[origin=c]{90}{\textbf{ChrF++}}}
&Llama-3-8B & 25.2 & 25.9 & 26.0 & 26.4 & 25.3 & 25.1 & 26.0 & 24.6 & 25.5 & 25.5 & 25.5 \\

&\quad + DPO &25.4&	26.3&	26.4&	26.5&	25.7&	25.4&	26.3&	24.8&	26.0&	25.9&	25.9\\
&\quad + SFT & 28.0 & 28.0 & 28.3 & 31.0 & 28.0 & 28.5 & 29.1 & 28.2 & 27.7 & 28.5 & 28.5 \\

&\quad + SFT + DPO & \textbf{30.57} & \textbf{29.6} & \textbf{29.9} & \textbf{32.4} & \textbf{29.4} & \textbf{30.2} & \textbf{30.7} & \textbf{29.8} & \textbf{29.3} & \textbf{29.6} & \textbf{30.0} \\
\cline{2-13}
&Gemma-3-12B & 27.1 & 26.3 & 27.8 & 26.7 & 26.7 & 26.5 & 26.9 & 26.9 & 27.2 & 26.3 & 26.8 \\
&\quad + DPO & 28.7 & 27.5 & 29.1 & 28.1 & 27.9 & 27.6 & 28.2 & 28.5 & 28.4 & 27.5 & 28.1 \\
&\quad + SFT & 34.8 & 33.0& 34.4 & 35.8 & 33.2 & 33.1 & 34.1 & 34.8 & 33.5 & 32.8 & 34.0 \\

&\quad + SFT + DPO & \textbf{36.1} & \textbf{35.1} & \textbf{36.6} & \textbf{37.6} & \textbf{35.4} & \textbf{35.9} & \textbf{36.1} & \textbf{36.1} & \textbf{35.7} & \textbf{34.7} & \textbf{35.9} \\
\midrule
\multirow{4}{*}{\rotatebox[origin=c]{90}{\textbf{LLM-Judge}}}
&Llama-3-8B & 53.0&	60.2&	58.6&	60.7&	59.5&	57.1&	61.0&	54.2&	59.5&	60.6&	58.5\\
&\quad+SFT+DPO &  \textbf{68.5} & \textbf{70.1} & \textbf{70.8} & \textbf{72.4} & \textbf{71.1} & \textbf{69.4} & \textbf{69.4} & \textbf{69.5} & \textbf{68.7} & \textbf{69.6} & \textbf{69.9} \\
\cline{2-13}
&Gemma-3-12B & 59.9&	60.8&	59.1&	60.0&	58.2&	60.6&	63.2&	59.4&	58.5&	60.0&	59.9 \\
&\quad+SFT+DPO & \textbf{63.2} & \textbf{66.9} & \textbf{68.3} & \textbf{66.8} & \textbf{64.1} & \textbf{63.9} & \textbf{65.2} & \textbf{65.7} & \textbf{66.6} & \textbf{68.2} & \textbf{65.9} \\
\bottomrule
\end{tabular}
}
\caption{\textsc{AfriLangTutor} evaluation results using ChrF++ as the automated evaluation metric. \textbf{LLM-as-a-judge} results average across the four judgment criteria with GPT-5.2 as the judge.} 
\label{tab:result1}
\end{table*}

\begin{figure}[t!]
    \centering
    \includegraphics[width=0.86\linewidth]{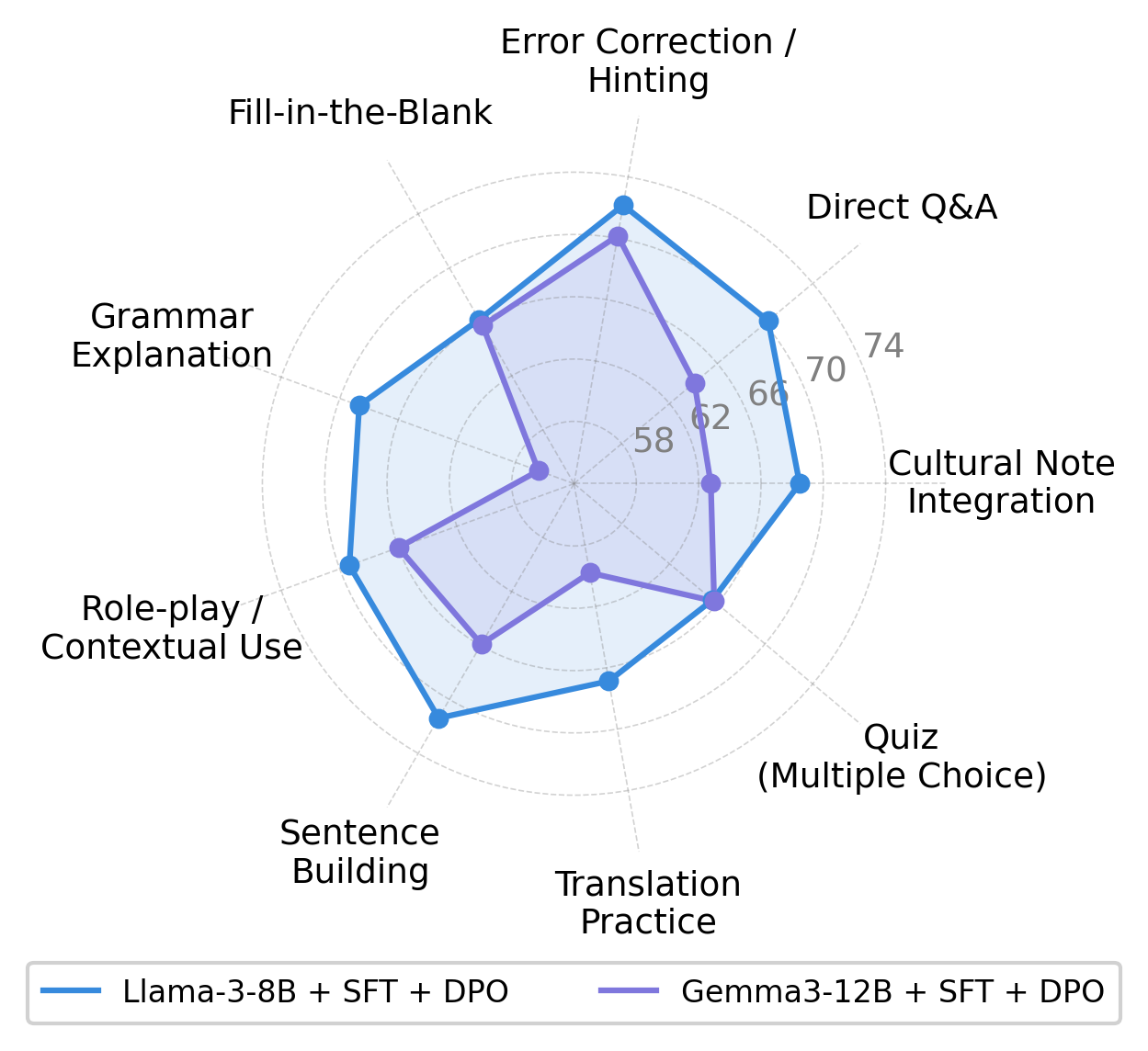}\vspace{-3mm}
    \caption{Performance of our \textsc{AfriLangTutor} LLMs (Llama-3-8B-IT and Gemma-3-12B-IT post SFT + DPO fine-tuning) across different question types in our \textsc{AfriLangEdu} benchmark test set.}
    \label{fig:dialog-performance-radar}\vspace{-4mm}
\end{figure}

\noindent\textbf{Baseline LLMs vs. SFT.} SFT yields a significant performance leap across all 10 languages. For example, in Gemma-3-12B-IT, the baseline average of 26.82 jumps to 33.97. This demonstrates that while instruction-tuned models have cross-lingual capabilities, they suffer from a \textit{low-resource gap} that general alignment cannot bridge. SFT with multi-turn dialogue data acts as a crucial domain adaptation step, teaching the model the specific structural nuances and tutoring style required for these languages.  \vspace{1.5mm}

\noindent\textbf{SFT vs. DPO vs. SFT + DPO.} Standard DPO (without a prior SFT phase) often underperforms compared to SFT alone. DPO is inherently a preference-alignment method, and not a knowledge-acquisition tool. When applied directly to a baseline that lacks sufficient language-specific grounding, the model struggles to distinguish \textit{better} answers because its base generation quality is already low. This confirms that for low-resource language tutoring, SFT is an important \textit{prerequisite} training step for meaningful preference learning. With performance averaged across languages post SFT + DPO, Llama-3-8B outperforms Gemma-3-12B across various query types while both models exhibit equivalent performance on \textit{fill-in-the-blank}, and \textit{multiple choice} queries, as shown in Figure \ref{fig:dialog-performance-radar}. Moreover, with performance averaged across all query sample types (Table \ref{tab:benchmark}) both models achieve similar results.

\section{Conclusion}

This paper presented a comprehensive framework for improving the tutoring capabilities of LLMs in low-resource African languages. We introduced \textsc{AfriLangDict}, a 194.7K multipurpose dictionary dataset covering 10 African languages; \textsc{AfriLangEdu}, a synthetic 78.9K multi-turn language tutoring dataset for SFT and DPO generated using \textsc{AfriLangDict} as seed data; and \textsc{AfriLangTutor}, multilingual LLMs fine-tuned from Llama-3-8B and Gemma-3-12B on \textsc{AfriLangEdu}. Using comprehensive evaluations with automated metrics such as ChrF++ and ROUGE-L, as well as GPT-5.2 as an LLM-as-a-judge, we demonstrated that \textsc{AfriLangTutor} LLMs achieve state-of-the-art performance compared to base models and other open-source LLMs 
AfriLangTutor models can be used and integrated into language tutoring platforms.


\section*{Limitations}
This work is not without limitations, and can be further extended in the following dimensions.

\paragraph{Multi-LLM data generation} While we study ten low-resource languages in this work, there are several others that also merit study and analysis in this context. Furthermore, while our automated dataset pipeline creates high-quality multilingual, multi-turn tutoring samples using dictionary seed data, it can be further refined using the latest multilingual LLMs and extensive human annotation. Moreover, while our method helps LLMs improve as LRL tutors in English, the pipeline is limited to dictionary-based generation settings. We defer the study of these limitations to future work.

\paragraph{Medium of instruction language} 
Since the current state of the art in large language models primarily targets English, most advancements and educational resources are also predominantly produced in English \cite{chu-etal-2025-llm,app15020671,fi17080366}. Accordingly, in this work, we used English as the medium of instruction when generating educational resources for low-resource languages. Our model is intended for English speakers who want to explore beginner-level new languages and cultures. As a next step, we plan to generate these educational resources directly in the target low-resource languages and evaluate the quality of LLM outputs. In the longer term, we aim to explore training multilingual tutoring models tailored to low-resource language contexts.

\paragraph{Advancing language tutoring}
While native speakers of the targeted African languages possess strong communicative competence, access to advanced linguistic education, including standardized language proficiency assessments and formal instruction, remains limited. The resources developed in this work, such as AfriLangEdu, provide a foundation to address this gap. Future work can build on these resources by generating more sophisticated linguistic data, incorporating expert-driven annotations, and developing advanced multilingual and multi-modal models tailored for high-level language tutoring and assessment.

\paragraph{Evaluation beyond in-house tests} Most existing evaluation datasets for the targeted languages focus on non-dialogue-based interactions and skills such as scientific or mathematical reasoning, which do not align with this work's objectives. There is no language learning-related test dataset available for AfriLangTutor models. Additionally, many of these benchmarks are not designed for generative tasks but rather for classical tasks such as classification and translation. Evaluate AfriLangTutor for other non-language-teaching tasks, which may be explored in the future.

\section*{Ethics Statement}
This work aims to support language learning and accessibility for low-resource languages that are underrepresented in NLP research. The \textsc{AfriLangDict} dataset is derived from various publicly available dictionary entries, ensuring no personal or sensitive data is included. While the automated generation of the \textsc{AfriLangEdu} dialogue dataset allows for scalable resource creation, we acknowledge the inherent risk of large language models introducing grammatical errors, hallucinations, or cultural inaccuracies, particularly in low-resource contexts. To mitigate this, we conducted a random-sampled manual evaluation with native speakers to verify the quality and appropriateness of the generated educational dialogues. The resulting datasets and \textsc{AfriLangTutor} models will be publicly released under open licenses to foster transparency and equitable access. We encourage future users to consider local linguistic nuances when deploying these models in real-world educational settings.

\bibliography{custom}

\appendix
\section*{Appendix}
\section{Data Generation Templates}\label{app:temlates}

Below are examples of generating templates with descriptions. We generate at least 3 full turns (3 from the language learner and the tutor's answers).

\begin{enumerate}
    \item \textbf{Direct Q\&A}: Simple student–tutor explanation of a word (phrase or sentence) meaning.
    \item \textbf{Quiz (Multiple Choice)}: language learner and the tutor interact in a question-and-answer conversation.
    \item \textbf{Fill-in-the-Blank}: A contextual sentence with a missing word.
    \item \textbf{Role-play / Contextual Use}: Greeting, school, or conversation simulation.
    \item \textbf{Error Correction / Hinting}: The student asks, and the tutor corrects the student’s misunderstanding.
    \item \textbf{Sentence Building}: The language learner (student) asks the tutor to build a sentence, and the tutor creates a complete sentence using the word.
    \item \textbf{Translation Practice}: Forward and backward translation check.
    \item \textbf{Spelling \& Pronunciation}: Language transliteration or phonetic spelling practice and spelling correction.
    \item \textbf{Cultural Note Integration}: Explanation of cultural or contextual relevance.
    \item \textbf{Grammar Explanation}: The student asks about a grammar rule involving the target word, and the tutor provides a clear and simple explanation with examples.
\end{enumerate}

For DPO training, by adding chosen and rejected features from the data, enabling the model to learn answering styles in various negative example scenarios from both the language learner (student) and language teacher (tutor) perspectives. Some examples of such cases for the DPO setting are presented below:
\begin{enumerate}
    \item \textbf{Misspelled / Typo}: The student attempts to ask about the target word but makes a significant spelling error (e.g., swapping letters, omitting vowels, or using phonetic spelling).
    \item \textbf{Vague / Ambiguous}: The student provides insufficient context or is unclear about their intent (e.g., typing only the word or asking "What about this?" without specifying the target).
    \item \textbf{Irrelevant / Mixed Context}: The student mixes the language-learning question with an unrelated topic (e.g., Python code, weather prediction, or general knowledge).
    \item \textbf{Factually Wrong Premise}: The student asks a question based on a confidently stated false assumption (e.g., "Since [WORD] means [WRONG\_MEANING], can I use it to describe a river?").
    \item \textbf{Out-of-Scope / Nonsensical}: The student asks for inappropriate usage (e.g., how to use the [WORD] as an insult) or poses impossible questions about abstract words (e.g., "What color is this verb?").
\end{enumerate}

\section{Influence Analysis}\label{app:influence-analysis}
We use TraceIn~\cite{pruthi2020estimatingtrainingdatainfluence} as the influence function for calculating the influence score between training and validation samples, thereby ensuring computational efficiency \cite{askari2025layerif, vitel2025first}. Let, the fine-tuned LLM parameterized by $\theta_L$. The influence score between a training sample $x^t_i$ and a validation sample $x^v_j$ can be defined as:

\begin{equation}
    \text{Influence}(x^t_i, x^v_j) = \nabla_{\theta_L} \ell(x^t_i, \theta_L) \cdot \nabla_{\theta_L} \ell(x^v_j, \theta_L)
\end{equation}

where $\ell(\cdot, \theta_L)$ is the cross-entropy loss.

\begin{figure}[h!]
    \centering
    \includegraphics[width=\linewidth]{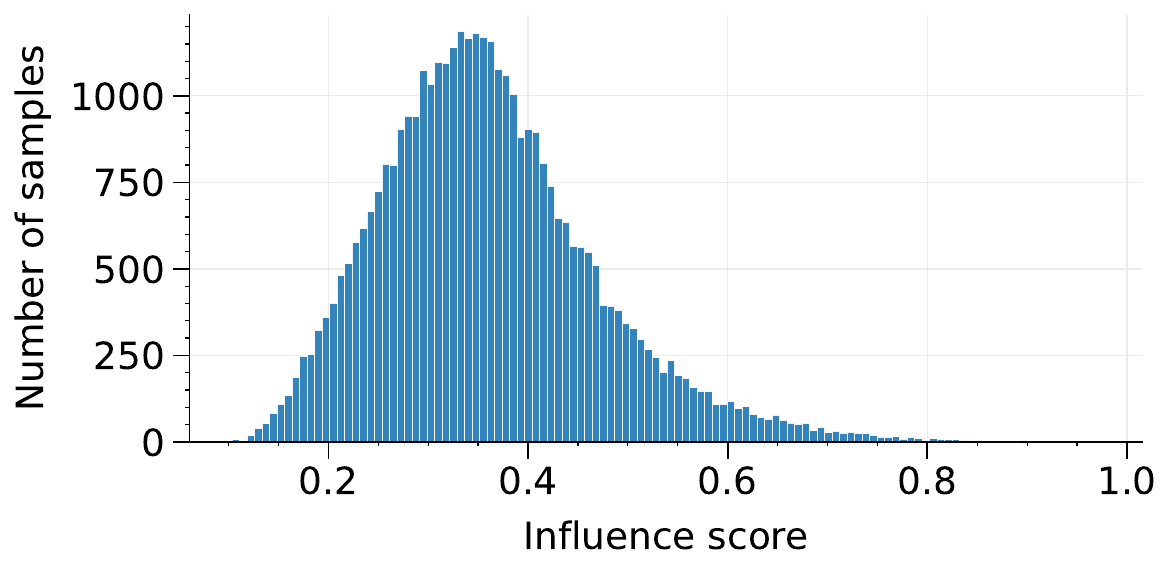}
    \caption{Average influence score distribution of the SFT training samples.}
    \label{fig:tracein}
\end{figure}

\looseness-1 Here, A positive influence score indicates that $x^t_i$ reduces the loss on $x^v_j$, suggesting a beneficial effect, while a negative score indicates a harmful influence. In our experiments, we found that most training samples have a positive influence. Figure \ref{fig:tracein} shows the distribution of the average influence scores of training samples over the validation set, illustrating that the majority of training samples contribute beneficially.

\section{LLM Judge Prompt Templates}\label{app:llm-judge-prompt}
Table \ref{tab:prompt} is the full prompt template for the LLM-as-a-judge evaluation across the four evaluation criteria (instructional alignment, pedagogical completeness, linguistic and cultural accuracy, and coherence and naturalness).

\begin{table*}[!h]
\begin{prompt}[label={box:nlp_kw}]{Orchid}{LLM Judge Prompt Template}
You are a skilled language tutoring evaluator tasked with judging the quality of a generated language teaching answer for a given query.
Instruction:
Score the answer generated by a system to a user's request on a Likert scale (1, 3, 5, 7) for four quality criteria.
Include a concise rationale for each score in less than 50 words (in English).
Rubric:
(1) Instruction Alignment\\
7: The response shows a perfect understanding of the user's intent. It follows all instructions, including task type (e.g., dialogue), formatting, and specific language constraints.\\
5: The response addresses the user's core question and follows most instructions, but may have minor slips in formatting or include slightly extraneous information.\\
3: The response only partially understands the user's intent. It might answer the question but fail to follow the requested format (e.g., providing a list when a dialogue was asked for) or omit a key constraint.\\
1: The response completely misses the user's intent, answers a different question, or fails to follow any of the structural instructions provided.\\
(2) Pedagogical Completeness\\
7: Provides a comprehensive learning "package": a clear explanation of the concept, multiple relevant examples, and a practice prompt or "scaffolding" that guides the learner to the next level.\\
5: Generally helpful for learning; provides an answer and an explanation, but the examples might be a bit generic or the scaffolding could be more structured.\\
3: Limited educational value. It might give the correct answer but offer a poor or overly brief explanation, leaving the learner without a clear understanding of the "why."\\
1: Pedagogically useless. Just gives the answer without any explanation or context, or provides misleading pedagogical advice that would confuse a student.\\
(3) Linguistic \& Cultural Accuracy\\
7: The language is flawless and sounds like a native speaker. It includes relevant cultural nuances (e.g., proper Amharic honorifics or social context) that make the language feel real and usable.\\
5: Grammatically correct with very minor slips. It is culturally acceptable but might rely on slightly formal or "textbook" phrasing that isn't commonly used in daily life.\\
3: Contains several grammatical or spelling errors. It may use literal translations that are technically "correct" but socially awkward or culturally weird in the target language.\\
1: Pervasive linguistic errors that distort meaning. The response is culturally insensitive or uses "hallucinated" grammar that would teach the learner incorrect habits.\\
(4) Coherence \& Naturalness\\
7: Perfectly organized with smooth transitions. The tone is encouraging, professional, and engaging. It reads like high-quality material from a professional language school.\\
5: Logical and easy to follow, though it may feel a bit mechanical. The tone is polite and professional, with minor gaps in the "flow" of conversation.\\
3: Noticeable issues in organization. The transitions between the explanation and the examples are clunky, or the tone feels inconsistent (e.g., shifting from overly formal to too casual).\\
1: Disjointed and difficult to follow. Ideas are disconnected, and the tone is either robotic, rude, or completely inappropriate for a tutoring environment.\\
Response Format (Strict JSON):\\
\{\{\\
  "instruction\_following\_rationale": "...", "instruction\_alignment\_score": SCORE,\\
  "pedagogical\_completeness\_rationale": "...", "pedagogical\_completeness\_score": SCORE,\\
  "linguistic\_cultural\_accuracy\_rationale": "...", "linguistic\_cultural\_accuracy\_score": SCORE,\\
  "coherence\_and\_naturalness\_rationale": "...", "coherence\_and\_naturalness\_score": SCORE\\
\}\}\\
Question: \{question\}\\
Candidate: \{candidate\}\\

\end{prompt}
\caption{Prompt details of the LLM-as-a-judge evaluation.}
\label{tab:prompt}
\end{table*}


\section{Human vs. LLM Judgment Agreement}\label{app:human-vs-llm}
\begin{figure*}[!h]
    \centering
    \includegraphics[width=1\linewidth]{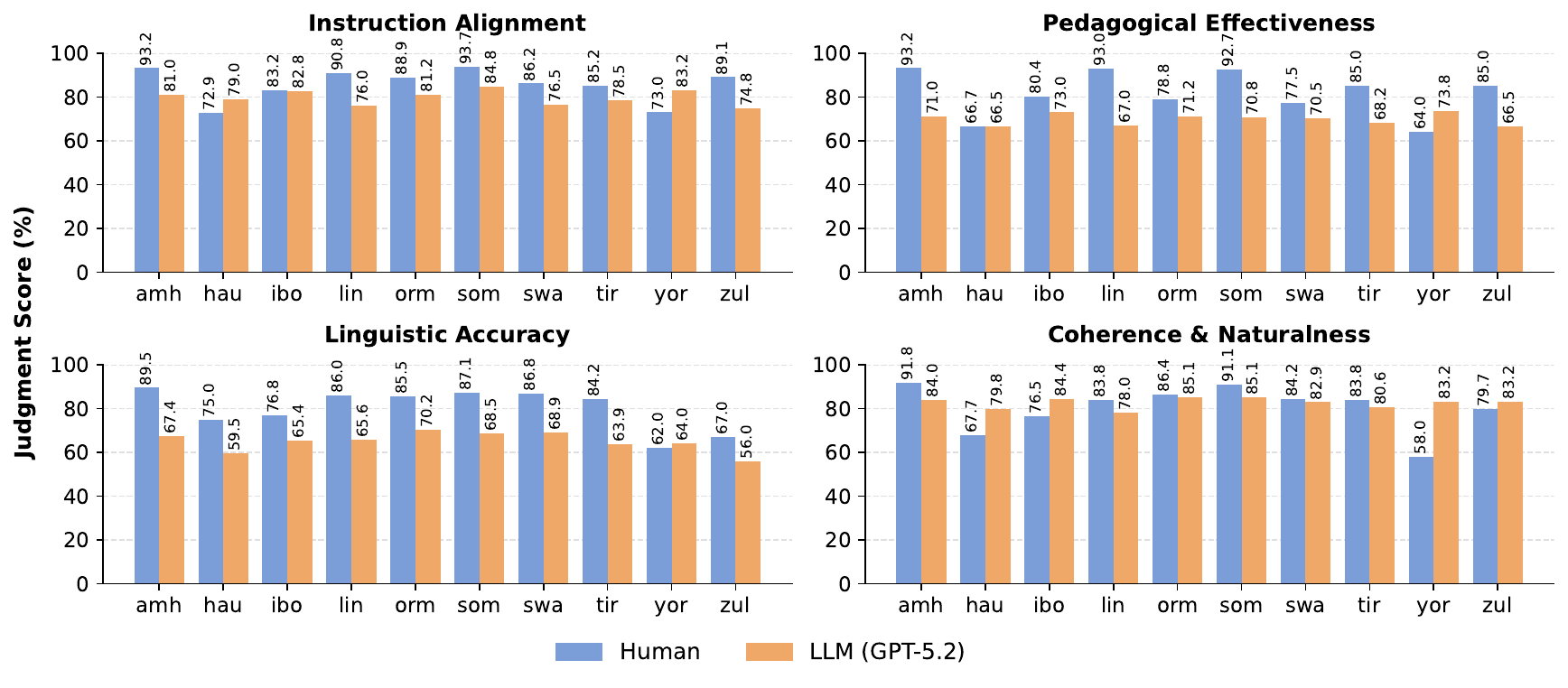}
    \caption{ Human vs LLMs (GPT-5.2) judgment agreement score (\%) across four judgment criteria, the 10 targeted languages, from $N=100$ samples in the test set.}
    \label{fig:hum-vs-llm-eval}
\end{figure*}

Figure \ref{fig:hum-vs-llm-eval} shows a comparison of Human  (native speakers) and LLM (GPT-5.2) evaluation scores across ten African languages using four evaluation dimensions on 100 samples. The score comparison reveals notable differences between Human and LLM-based evaluation across languages. Human evaluators consistently assign higher scores than the LLM in most evaluation dimensions, especially in Pedagogical Effectiveness and Linguistic Accuracy, suggesting that the LLM may underestimate pedagogical quality and cultural-linguistic nuances in multilingual educational responses. In contrast, the gap between Human and LLM evaluations is comparatively smaller for Coherence and Naturalness, indicating stronger agreement on fluency-related aspects. Language-specific variations are also evident, where languages such as Amharic (amh), Somali (som), and Oromo (orm) achieve relatively high Human scores across all dimensions, while languages such as Yoruba (yor) and Zulu (zul) exhibit more discrepancies between Human and LLM assessments. Overall, the results highlight both the potential and limitations of LLM-based automatic evaluation for culturally diverse languages.

\section{Additional Ablation Results} \label{app:addional-results}
Table \ref{tab:result2} shows additional ChrF++ results from different fine-tuning settings. 
\paragraph{LoRA vs. Full SFT} Full SFT (28.93) slightly outperforms SFT with LoRA (28.48) in the Llama-3 experiments. While LoRA is parameter-efficient and prevents catastrophic forgetting, Full SFT allows the model to undergo more significant representational shifts. In low-resource scenarios where the model needs to learn a new script or complex morphology, the additional degrees of freedom in full SFT provide a marginal but consistent advantage.

\paragraph{DPO vs. Self-DPO} Self-DPO (where the rejected answer is from the model itself) underperforms compared to using an external rejected model. For Llama + SFT, DPO$^2$ (the rejected is from Gemma-2-2b-it) is significantly better than the result from self DPO. This suggests that in low-resource contexts, "self-critique" is limited by the model's own lack of knowledge. Using a diverse set of rejected answers from other LLMs (like Gemma-2-2b) provides a clearer contrastive signal, helping the model identify and avoid a wider variety of linguistic errors.

\begin{table*}[h!]
\centering
\caption{ChrF++ results with different settings. For the DPO experiment, the chosen answer is from Gemini 2.5 pro and the rejected answer is generated from Llama-3.1-8B-Instruct$^1$, Gemma-2-2b-it$^2$, and \textbf{self} DPO (the DPO rejected answer is generated from Llama-3-8B-IT itself). Param$^1$ = $\beta$ 0.1, batch size 1, epoch 3, and Param$^2$ = $\beta$ 0.5, batch size 4, epoch 10. SFT is with LORA, and FULL SFT is full fine-tuning.}

\resizebox{\textwidth}{!}{
\begin{tabular}{rcccccccccc|c}
\toprule
\textbf{Model} & \textbf{amh} &\textbf{hau}& \textbf{ibo}& \textbf{lin}& \textbf{orm}& \textbf{som}& \textbf{swa}& \textbf{tir}& \textbf{yor}& \textbf{zul}& \textbf{Avg}\\
\hline
Llama-3-8B-IT & 25.17 & 25.87 & 25.96 & 26.35 & 25.33 & 25.05 & 25.97 & 24.56 & 25.48 & 25.51 & 25.52 \\
\quad + DPO$^1$ & 21.41 & 21.80 & 22.77 & 22.28 & 19.74 & 20.94 & 22.21 & 20.01 & 21.86 & 19.78 & 21.29 \\
\quad + DPO$^2$ & 28.37 & 29.88 & 30.20 & 30.67 & 29.82 & 29.43 & 30.48 & 27.95 & 30.01 & 29.05 & 29.59 \\
\quad + DPO$^1$ + Param$^1$ &21.41&	21.80&	22.77&	22.28&	19.74&	20.94&	22.21&	20.01&	21.86&	19.78&	21.29\\
\quad + DPO$^1$ + Param$^2$ &24.90&	25.61&	25.69&	25.87&	24.93&	24.81&	25.59&	24.02&	24.97&	25.15&	25.15\\

\quad + DPO$^2$ + Param$^1$ & 28.37 & 29.88 & 30.20 & 30.67 & 29.82 & 29.43 & 30.48 & 27.95 & 30.01 & 29.05 & 29.59\\
\quad + DPO$^2$ + Param$^2$ &25.40&	26.29&	26.41&	26.50&	25.70&	25.40&	26.30&	24.81&	25.96&	25.93&	25.87\\
\quad + SFT & 28.03 & 27.98 & 28.25 & 30.98 & 27.98 & 28.54 & 29.05 & 28.17 & 27.66 & 28.51 & 28.48 \\
\quad + SFT (influence 90\%) & 28.20&	27.49&	28.56&	28.25&	28.24&	30.70&	27.69&	28.72&	27.99&	28.97&	28.45 \\

\quad + FULL SFT & 28.45 & 28.32 & 28.94 & 31.72 & 28.56 & 28.90 & 28.87 & 28.79 & 28.39 & 28.66 & 28.93 \\
\quad + SFT + DPO\textsuperscript{self} & 24.91&	25.70&	25.47&	27.21&	24.93&	25.75&	25.88&	25.03&	24.78&	25.56&	25.5\\
\quad + FULL SFT + DPO\textsuperscript{self} & 23.27&	23.55&	23.76&	26.30&	23.38&	23.90&	24.36&	23.59&	23.00&	23.50&	23.83 \\
\quad + SFT + DPO$^1$ & 28.05 & 27.99 & 27.89 & 30.61 & 27.79 & 28.54 & 28.82 & 28.12 & 27.46 & 28.19 & 28.32 \\
\quad + SFT + DPO$^2$ & 29.57 & 29.60 & 29.90 & 32.41 & 29.37 & 30.21 & 30.69 & 29.80 & 29.33 & 29.62 & \textbf{30.02} \\
\midrule
Gemma-3-12B-IT & 27.05 & 26.27 & 27.76 & 26.68 & 26.65 & 26.45 & 26.86 & 26.93 & 27.19 & 26.27 & 26.82 \\
\quad + DPO$^1$ & 27.50 & 26.79 & 28.20 & 27.02 & 27.05 & 26.71 & 27.11 & 27.46 & 27.53 & 26.67 & 27.22 \\
\quad + DPO$^2$ & 28.65 & 27.50 & 29.05 & 28.06 & 27.88 & 27.60 & 28.15 & 28.52 & 28.38 & 27.49 & 28.14 \\
\quad + SFT & 34.84 & 32.99 & 34.43 & 35.82 & 33.23 & 33.15 & 34.08 & 34.75 & 33.45 & 32.80 & 33.97 \\
\quad + SFT (influence 90\%) & 34.95&	33.78&	34.68&	33.18&	34.55&	35.94&	31.96&	33.37&	33.54&	34.22&	34.02 \\

\quad + SFT + DPO$^1$ & 35.52 & 33.53 & 35.43 & 36.22 & 34.08 & 34.57 & 35.06 & 35.41 & 34.79 & 33.72 & 34.85 \\
\quad + SFT + DPO$^2$ & \textbf{36.08} & \textbf{35.02} & \textbf{36.62} & \textbf{37.55} & \textbf{35.42} & \textbf{35.87} & \textbf{36.06} & \textbf{36.01} & \textbf{35.67} & \textbf{34.71} & \textbf{35.90} \\
\bottomrule
\end{tabular}
}
\label{tab:result2}
\end{table*}

\section{Benchmark Result Details from LLM-as-a-judge Evaluation}\label{app:benchmark-details}
Table \ref{tab:llm-benchmark-detail} shows the benchmark results from the LLM-as-a-judge Evaluation. The rubric scaling results are in converted \% from 100 for the ease of understanding.

\begin{table*}[h!]
\centering
\caption{LLM Judge detail results from GPT 5.2. The results are the Rubric criteria of LLM judgment converted to 100\%. The details of the instruction templates are in Appendix \ref{app:llm-judge-prompt}.}

\resizebox{\textwidth}{!}{
\begin{tabular}{llcccccccccc|c}
\toprule
\textbf{Model} &\textbf{Judge Criteria}& \textbf{amh} &\textbf{hau}& \textbf{ibo}& \textbf{lin}& \textbf{orm}& \textbf{som}& \textbf{swa}& \textbf{tir}& \textbf{yor}& \textbf{zul}& \textbf{Avg}\\
\hline
\multirow{4}{*}{{Llama-3.2-1B-IT}} 
&Instruction alignment & 57.6 & 64.7 & 66.8 & 66.9 & 67.5 & 66.0 & 64.9 & 59.2 & 65.3 & 67.1 & 64.6 \\
&Pedagogical completeness & 38.7 & 45.8 & 43.4 & 42.0 & 43.8 & 41.6 & 42.2 & 40.3 & 43.4 & 44.2 & 42.5 \\
&Linguistic \& Cultural Accuracy & 28.1 & 34.9 & 33.8 & 35.9 & 38.5 & 34.7 & 33.8 & 28.8 & 35.2 & 37.9 & 34.2 \\
&Coherence \& Naturalness & 54.4 & 65.0 & 61.9 & 62.1 & 65.4 & 61.8 & 60.5 & 56.6 & 62.2 & 65.1 & 61.5 \\
\cdashline{3-13}
& & 44.7&	52.6&	51.48&	51.7&	53.8&	51.0&	50.4&	46.2&	51.5&	53.6&	50.7\\
\hline

\multirow{4}{*}{{Gemma-2-2b-it}} 
&Instruction alignment & 63.2 & 69.2 & 66.0 & 68.5 & 68.3 & 66.7 & 69.6 & 64.6 & 68.8 & 68.8 & 67.4 \\
&Pedagogical completeness & 48.4 & 54.6 & 52.6 & 52.7 & 52.2 & 51.3 & 55.2 & 49.5 & 53.0 & 53.0 & 52.3 \\
&Linguistic \& Cultural Accuracy & 34.4 & 44.1 & 43.0 & 43.4 & 45.4 & 41.7 & 44.7 & 36.0 & 44.3 & 49.3 & 42.7 \\
&Coherence \& Naturalness & 59.0 & 69.2 & 63.2 & 68.6 & 69.0 & 64.9 & 68.8 & 60.0 & 65.4 & 70.8 & 65.9 \\
\cdashline{3-13}
& & 51.3&	59.3&	56.2&	58.3&	58.7&	56.2&	59.6&	52.5&	57.9&	60.5&	57.1 \\
\hline

\multirow{4}{*}{{Llama-3.2-3B-IT}} 
&Instruction alignment & 58.3 & 69.8 & 70.5 & 71.7 & 68.3 & 67.9 & 71.4 & 60.2 & 68.2 & 70.7 & 67.7 \\
&Pedagogical completeness & 35.0 & 44.6 & 45.9 & 43.6 & 40.9 & 42.1 & 46.7 & 35.4 & 43.9 & 44.4 & 42.3 \\
&Linguistic \& Cultural Accuracy & 39.3 & 54.8 & 43.5 & 49.1 & 54.9 & 45.2 & 49.2 & 42.9 & 46.5 & 56.7 & 48.2 \\
&Coherence \& Naturalness & 57.2 & 70.8 & 66.4 & 68.5 & 69.4 & 64.6 & 67.3 & 59.4 & 65.9 & 70.8 & 66.0 \\
\cdashline{3-13}
& & 47.5&	60.0&	56.6&	58.2&	58.4&	55.0&	58.7&	49.5&	56.1&	60.7&	56.1\\
\hline

\multirow{4}{*}{{Aya-global-3.5B}} 
&Instruction alignment & 67.9 & 70.6 & 71.6 & 69.6 & 67.8 & 67.5 & 71.1 & 66.7 & 69.7 & 69.8 & 69.2 \\
&Pedagogical completeness & 52.7 & 51.0 & 55.6 & 50.4 & 44.5 & 47.8 & 54.2 & 52.1 & 49.9 & 50.0 & 50.8 \\
&Linguistic \& Cultural Accuracy & 40.7 & 41.2 & 45.2 & 38.4 & 39.2 & 36.0 & 44.6 & 38.8 & 44.1 & 41.4 & 41.0 \\
&Coherence \& Naturalness & 62.4 & 66.1 & 67.4 & 66.2 & 63.5 & 62.6 & 68.4 & 63.1 & 65.4 & 66.4 & 65.2 \\
\cdashline{3-13}
& & 55.9&	57.2&	60.0&	56.2&	53.8&	53.5&	59.6&	55.2&	57.3&	56.9&	56.6 \\
\hline

\multirow{4}{*}{{Gemma-3-4b-IT}} 
&Instruction alignment & 66.1 & 67.7 & 67.3 & 69.2 & 67.4 & 68.4 & 69.3 & 66.0 & 67.2 & 68.3 & 67.7 \\
&Pedagogical completeness & 53.1 & 54.5 & 53.2 & 55.4 & 53.3 & 53.3 & 56.4 & 53.4 & 53.5 & 54.2 & 54.0 \\
&Linguistic \& Cultural Accuracy & 38.4 & 38.4 & 38.9 & 39.0 & 36.8 & 36.7 & 40.3 & 39.4 & 38.7 & 36.6 & 38.3 \\
&Coherence \& Naturalness & 57.5 & 60.4 & 57.7 & 62.7 & 60.5 & 59.4 & 62.5 & 58.0 & 57.9 & 61.4 & 59.8 \\
\cdashline{3-13}
& & 53.8&	55.3&	54.3&	56.6&	54.5&	54.5&	57.1&	54.2&	54.3&	55.1&	55.0 \\
\hline

\multirow{4}{*}{{Qwen3-4B-IT}} 
&Instruction alignment & 60.3 & 65.8 & 65.5 & 68.3 & 66.0 & 65.5 & 66.5 & 60.7 & 65.6 & 66.7 & 65.1 \\
&Pedagogical completeness & 46.9 & 51.7 & 52.1 & 51.4 & 51.9 & 50.6 & 49.9 & 47.6 & 50.6 & 51.0 & 50.4 \\
&Linguistic \& Cultural Accuracy & 33.8 & 49.5 & 43.0 & 42.9 & 48.6 & 43.7 & 42.9 & 34.2 & 45.5 & 49.1 & 43.3 \\
&Coherence \& Naturalness & 51.4 & 61.3 & 59.3 & 60.8 & 63.2 & 59.1 & 58.1 & 51.5 & 58.6 & 62.7 & 58.6 \\
\cdashline{3-13}
& & 48.1&	57.1&	55.0&	55.9&	57.4&	54.7&	54.4&	48.5&	55.1&	57.4&	54.4\\
\hline

\multirow{4}{*}{{Llama-3-8B-IT}} 
&Instruction alignment & 68.2 & 70.8 & 72.3 & 74.6 & 72.1 & 71.1 & 73.3 & 69.7 & 71.8 & 72.2 & 71.6 \\
&Pedagogical completeness & 48.8 & 51.4 & 52.2 & 51.6 & 48.5 & 49.3 & 53.2 & 50.0 & 51.7 & 50.8 & 50.8 \\
&Linguistic \& Cultural Accuracy & 32.7 & 47.7 & 41.5 & 45.2 & 46.8 & 40.5 & 47.3 & 33.3 & 45.2 & 47.7 & 42.8 \\
&Coherence \& Naturalness & 62.4 & 70.8 & 68.5 & 71.5 & 70.6 & 67.5 & 70.3 & 63.6 & 69.2 & 71.6 & 68.6 \\
\cdashline{3-13}
& & 53.0&	60.2&	58.6&	60.7&	59.5&	57.1&	61.0&	54.2&	59.5&	60.6& 58.5\\
\hline

\multirow{4}{*}{{Gemma-3-12B-IT}} 
&Instruction alignment & 69.7 & 72.1 & 69.0 & 71.8 & 68.2 & 71.2 & 73.2 & 69.6 & 68.8 & 71.6 & 70.5 \\
&Pedagogical completeness & 57.4 & 59.0 & 56.8 & 57.9 & 56.4 & 58.8 & 60.9 & 57.0 & 56.9 & 58.3 & 57.9 \\
&Linguistic \& Cultural Accuracy & 49.9 & 47.2 & 50.2 & 45.3 & 44.4 & 49.0 & 52.5 & 49.0 & 47.4 & 45.5 & 48.0 \\
&Coherence \& Naturalness & 62.4 & 64.9 & 60.4 & 65.1 & 63.7 & 63.3 & 66.2 & 62.0 & 60.9 & 64.5 & 63.2 \\
\cdashline{3-13}
& & 59.9&	60.8&	59.1&	60.0&	58.2&	60.6&	63.2&	59.4&	58.5&	60.0&	59.9 \\

\hline
\multirow{4}{*}{{AfriqueGemma-4B}} 
&Instruction alignment & 38.8 & 40.9 & 36.8 & 37.4 & 39.9 & 37.0 & 38.2 & 39.1 & 36.7 & 39.9 & 38.5 \\
&Pedagogical completeness & 28.7 & 29.6 & 28.8 & 27.4 & 28.9 & 28.1 & 28.5 & 28.7 & 28.4 & 28.7 & 28.6 \\
&Linguistic \& Cultural Accuracy & 36.9 & 34.3 & 33.8 & 32.2 & 35.7 & 34.7 & 35.0 & 36.5 & 32.9 & 34.8 & 34.7 \\
&Coherence \& Naturalness & 28.9 & 29.7 & 28.7 & 27.6 & 29.2 & 28.4 & 29.0 & 28.5 & 28.6 & 28.9 & 28.7 \\
\cdashline{3-13}
& & 33.3 & 33.6 & 32.0 & 31.2 & 33.4 & 32.1 & 32.7 & 33.2 & 31.7 & 33.1 & 32.6 \\
\hline
\multirow{4}{*}{{AfriqueLlama-8B}} 
&Instruction alignment & 42.6 & 42.7 & 37.0 & 37.0 & 40.4 & 39.5 & 40.5 & 42.4 & 37.0 & 38.7 & 39.8 \\
&Pedagogical completeness & 30.3 & 31.9 & 29.0 & 28.4 & 30.3 & 29.6 & 30.2 & 30.5 & 29.1 & 29.4 & 29.9 \\
&Linguistic \& Cultural Accuracy & 39.1 & 34.3 & 33.1 & 31.5 & 35.2 & 34.7 & 36.6 & 39.2 & 33.3 & 32.9 & 35.0 \\
&Coherence \& Naturalness & 30.6 & 30.3 & 28.8 & 27.8 & 29.1 & 28.7 & 29.3 & 30.8 & 28.6 & 28.8 & 29.3 \\
\cdashline{3-13}
& & 35.7 & 34.8 & 32.0 & 31.2 & 33.8 & 33.1 & 34.2 & 35.7 & 32.0 & 32.5 & 33.5 \\
\hline
\multirow{4}{*}{{AfriqueGemma-12B}} 
&Instruction alignment & 34.1 & 36.7 & 36.6 & 33.7 & 34.6 & 35.0 & 36.4 & 35.3 & 35.6 & 35.8 & 35.4 \\
&Pedagogical completeness & 28.6 & 30.0 & 29.6 & 27.7 & 28.7 & 29.0 & 29.4 & 29.4 & 29.3 & 28.9 & 29.1 \\
&Linguistic \& Cultural Accuracy & 34.2 & 34.2 & 37.1 & 33.0 & 34.6 & 34.5 & 36.8 & 34.9 & 35.4 & 35.1 & 35.0 \\
&Coherence \& Naturalness & 28.8 & 30.9 & 29.8 & 28.7 & 28.9 & 29.2 & 30.0 & 29.2 & 29.7 & 29.9 & 29.5 \\
\cdashline{3-13}
& & 31.4 & 33.0 & 33.3 & 30.8 & 31.7 & 31.9 & 33.2 & 32.2 & 32.5 & 32.4 & 32.3 \\
\hline
\multirow{4}{*}{{AfriqueQwen-14B}} 
&Instruction alignment & 50.3 & 48.3 & 52.5 & 48.3 & 47.2 & 48.3 & 45.2 & 51.0 & 48.4 & 46.2 & 48.6 \\
&Pedagogical completeness & 36.1 & 34.8 & 36.9 & 33.5 & 34.7 & 35.0 & 34.2 & 36.2 & 35.3 & 34.1 & 35.1 \\
&Linguistic \& Cultural Accuracy & 39.4 & 35.6 & 39.5 & 34.4 & 37.2 & 37.5 & 35.5 & 39.5 & 36.4 & 34.3 & 36.9 \\
&Coherence \& Naturalness & 40.0 & 41.9 & 45.4 & 40.5 & 40.8 & 41.7 & 40.1 & 39.2 & 42.5 & 40.3 & 41.2 \\
\cdashline{3-13}
& & 41.5 & 40.2 & 43.6 & 39.2 & 40.0 & 40.6 & 38.8 & 41.5 & 40.7 & 38.7 & 40.5 \\
\bottomrule
\end{tabular}
}

\label{tab:llm-benchmark-detail}
\end{table*}

\section{Benchmark Result Details (rating)}\label{app:benchmark-ratings}
Table \ref{tab:benckmark-ratings} shows the LLM-judge rating result details across benchmarked LLMs.

\begin{table*}[!h]
\centering
\caption{LLM Judge rating Results by Language and LLM. The rating classes are in the Rubric across the four judging criteria (1. Instruction Alignment Score, 2. Pedagogical Completeness Score, 3. Linguistic Cultural Accuracy Score, and 4. Coherence and Naturalness Score), and the rating values are either 1, 3, 5, or 7. The average for the corresponding languages is also presented at the end of each LLM (below the dashed line).}

\resizebox{\textwidth}{!}
{
\begin{tabular}{llcccccccccc|c}
\toprule
\textbf{Model} &\textbf{Judge Criteria}& \textbf{amh} &\textbf{hau}& \textbf{ibo}& \textbf{lin}& \textbf{orm}& \textbf{som}& \textbf{swa}& \textbf{tir}& \textbf{yor}& \textbf{zul}& \textbf{Avg}\\
\hline
\multirow{4}{*}{Llama-3.2-1B-IT} & Instruction Alignment & 3.61 & 4.17 & 4.34 & 4.35 & 4.40 & 4.28 & 4.19 & 3.74 & 4.23 & 4.37 & 4.17 \\
 & Pedagogical Completeness & 2.09 & 2.66 & 2.47 & 2.36 & 2.50 & 2.33 & 2.37 & 2.23 & 2.47 & 2.53 & 2.40 \\
 & Linguistic Cultural Accuracy & 1.25 & 1.79 & 1.70 & 1.88 & 2.08 & 1.78 & 1.70 & 1.30 & 1.81 & 2.03 & 1.73 \\
 & Coherence And Naturalness & 3.35 & 4.20 & 3.95 & 3.96 & 4.24 & 3.94 & 3.84 & 3.53 & 3.97 & 4.21 & 3.92 \\
\cdashline{3-13}
&& 2.56&	3.21&	3.12&	3.14&	3.31&	3.08&	3.03&	2.70&	3.12&	3.29&	3.06\\
\hline
\multirow{4}{*}{Gemma-2-2b-it} & Instruction Alignment & 4.06 & 4.54 & 4.28 & 4.48 & 4.47 & 4.34 & 4.57 & 4.17 & 4.50 & 4.50 & 4.39 \\
 & Pedagogical Completeness & 2.87 & 3.37 & 3.20 & 3.22 & 3.18 & 3.10 & 3.42 & 2.96 & 3.24 & 3.24 & 3.18 \\
 & Linguistic Cultural Accuracy & 1.75 & 2.52 & 2.44 & 2.47 & 2.63 & 2.34 & 2.58 & 1.88 & 2.54 & 2.95 & 2.41 \\
 & Coherence And Naturalness & 3.72 & 4.54 & 4.06 & 4.49 & 4.52 & 4.20 & 4.50 & 3.80 & 4.23 & 4.66 & 4.27 \\
\cdashline{3-13}
&& 3.10&	3.74&	3.50&	3.67&	3.70&	3.50&	3.77&	3.20&	3.63&	3.84&	3.56\\
\hline
\multirow{4}{*}{Llama-3.2-3B-IT} & Instruction Alignment Score & 3.66 & 4.59 & 4.64 & 4.74 & 4.47 & 4.43 & 4.71 & 3.82 & 4.45 & 4.65 & 4.42 \\
 & Pedagogical Completeness Score & 1.80 & 2.57 & 2.67 & 2.49 & 2.28 & 2.37 & 2.74 & 1.83 & 2.51 & 2.55 & 2.38 \\
 & Linguistic Cultural Accuracy Score & 2.14 & 3.39 & 2.48 & 2.93 & 3.39 & 2.62 & 2.94 & 2.43 & 2.72 & 3.54 & 2.86 \\
 & Coherence And Naturalness Score & 3.58 & 4.66 & 4.31 & 4.48 & 4.55 & 4.17 & 4.38 & 3.75 & 4.27 & 4.66 & 4.28 \\
 \cdashline{3-13}
&& 2.80&	3.80&	3.53&	3.66&	3.67&	3.40&	3.69&	2.96&	3.49&	3.85&	3.49\\
\hline
\multirow{4}{*}{Aya-global-3.5B} & Instruction Alignment Score & 4.43 & 4.65 & 4.73 & 4.57 & 4.42 & 4.40 & 4.68 & 4.34 & 4.58 & 4.59 & 4.54 \\
 & Pedagogical Completeness Score & 3.21 & 3.08 & 3.45 & 3.03 & 2.56 & 2.82 & 3.34 & 3.17 & 2.99 & 3.00 & 3.07 \\
 & Linguistic Cultural Accuracy Score & 2.25 & 2.30 & 2.62 & 2.07 & 2.14 & 1.88 & 2.57 & 2.10 & 2.53 & 2.31 & 2.28 \\
 & Coherence And Naturalness Score & 3.99 & 4.29 & 4.39 & 4.30 & 4.08 & 4.01 & 4.47 & 4.05 & 4.23 & 4.32 & 4.21 \\
  \cdashline{3-13}
&& 3.47&	3.58&	3.80&	3.49&	3.30&	3.28&	3.77&	3.42&	3.58&	3.56&	3.53\\
\hline
\multirow{4}{*}{Gemma-3-4b-IT} & Instruction Alignment Score & 4.29 & 4.41 & 4.39 & 4.53 & 4.39 & 4.47 & 4.55 & 4.28 & 4.38 & 4.46 & 4.42 \\
 & Pedagogical Completeness Score & 3.25 & 3.36 & 3.26 & 3.43 & 3.27 & 3.26 & 3.51 & 3.27 & 3.28 & 3.33 & 3.32 \\
 & Linguistic Cultural Accuracy Score & 2.07 & 2.07 & 2.11 & 2.12 & 1.95 & 1.94 & 2.22 & 2.16 & 2.10 & 1.93 & 2.07 \\
 & Coherence And Naturalness Score & 3.60 & 3.83 & 3.61 & 4.01 & 3.84 & 3.75 & 4.00 & 3.64 & 3.63 & 3.91 & 3.79 \\
   \cdashline{3-13}
&& 3.30&	3.42&	3.34&	3.52&	3.36&	3.36&	3.57&	3.34&	3.35&	3.41&	3.40\\
\hline
\multirow{4}{*}{Qwen3-4B-IT} & Instruction Alignment Score & 3.83 & 4.26 & 4.24 & 4.46 & 4.28 & 4.24 & 4.32 & 3.85 & 4.25 & 4.34 & 4.21 \\
 & Pedagogical Completeness Score & 2.75 & 3.13 & 3.17 & 3.11 & 3.15 & 3.05 & 2.99 & 2.81 & 3.05 & 3.08 & 3.03 \\
 & Linguistic Cultural Accuracy Score & 1.70 & 2.96 & 2.44 & 2.43 & 2.88 & 2.50 & 2.43 & 1.74 & 2.64 & 2.93 & 2.47 \\
 & Coherence And Naturalness Score & 3.12 & 3.90 & 3.74 & 3.86 & 4.05 & 3.72 & 3.65 & 3.12 & 3.69 & 4.02 & 3.69 \\
\cdashline{3-13}
&& 2.85&	3.56&	3.40&	3.47&	3.59&	3.38&	3.35&	2.88&	3.41&	3.59&	3.35\\
\hline
\multirow{4}{*}{Llama-3-8B-IT} & Instruction Alignment Score & 4.45 & 4.67 & 4.78 & 4.97 & 4.77 & 4.69 & 4.86 & 4.58 & 4.75 & 4.78 & 4.73 \\
 & Pedagogical Completeness Score & 2.91 & 3.11 & 3.18 & 3.13 & 2.88 & 2.94 & 3.25 & 3.00 & 3.14 & 3.06 & 3.06 \\
 & Linguistic Cultural Accuracy Score & 1.61 & 2.82 & 2.32 & 2.62 & 2.74 & 2.24 & 2.78 & 1.67 & 2.62 & 2.82 & 2.42 \\
 & Coherence And Naturalness Score & 3.99 & 4.67 & 4.48 & 4.72 & 4.65 & 4.40 & 4.62 & 4.09 & 4.54 & 4.73 & 4.49 \\
 \cdashline{3-13}
&& 3.24&	3.82&	3.69&	3.86&	3.76&	3.57&	3.88&	3.34&	3.76&	3.85&	3.68\\
\hline
\multirow{4}{*}{Gemma-3-12B-IT} 
& Instruction Alignment Score & 4.58 & 4.76 & 4.52 & 4.75 & 4.45 & 4.70 & 4.86 & 4.57 & 4.51 & 4.73 & 4.64 \\
 & Pedagogical Completeness Score & 3.59 & 3.72 & 3.54 & 3.63 & 3.51 & 3.70 & 3.87 & 3.56 & 3.55 & 3.66 & 3.64 \\
 & Linguistic Cultural Accuracy Score & 2.99 & 2.77 & 3.01 & 2.62 & 2.55 & 2.92 & 3.20 & 2.92 & 2.80 & 2.64 & 2.84 \\
 & Coherence And Naturalness Score & 3.90 & 4.19 & 3.83 & 4.21 & 4.09 & 4.06 & 4.30 & 3.96 & 3.87 & 4.16 & 4.06 \\
\cdashline{3-13}
&& 3.77&	3.86&	3.73&	3.80&	3.65&	3.85&	4.06&	3.75&	3.68&	3.80&	3.80\\
\hline
\multirow{4}{*}{AfriqueGemma-4B} 
& Instruction Alignment Score & 2.11 & 2.28 & 1.94 & 1.99 & 2.19 & 1.96 & 2.06 & 2.13 & 1.94 & 2.19 & 2.08 \\
 & Pedagogical Completeness Score & 1.29 & 1.37 & 1.30 & 1.19 & 1.32 & 1.25 & 1.28 & 1.30 & 1.27 & 1.29 & 1.29 \\
 & Linguistic Cultural Accuracy Score & 1.95 & 1.74 & 1.70 & 1.58 & 1.85 & 1.77 & 1.80 & 1.92 & 1.63 & 1.78 & 1.77 \\
 & Coherence And Naturalness Score & 1.32 & 1.37 & 1.29 & 1.21 & 1.33 & 1.27 & 1.32 & 1.28 & 1.28 & 1.31 & 1.30 \\
\cdashline{3-13}
&& 1.67 & 1.67 & 1.56 & 1.49 & 1.67 & 1.56 & 1.61 & 1.66 & 1.53 & 1.64 & 1.66 \\
\hline
\multirow{4}{*}{AfriqueLlama-8B} 
& Instruction Alignment Score & 2.41 & 2.42 & 1.96 & 1.96 & 2.23 & 2.16 & 2.24 & 2.39 & 1.96 & 2.10 & 2.18 \\
 & Pedagogical Completeness Score & 1.42 & 1.55 & 1.32 & 1.27 & 1.42 & 1.36 & 1.42 & 1.44 & 1.33 & 1.36 & 1.39 \\
 & Linguistic Cultural Accuracy Score & 2.13 & 1.74 & 1.65 & 1.52 & 1.82 & 1.78 & 1.93 & 2.14 & 1.66 & 1.63 & 1.80 \\
 & Coherence And Naturalness Score & 1.45 & 1.42 & 1.30 & 1.22 & 1.33 & 1.30 & 1.35 & 1.46 & 1.29 & 1.30 & 1.34 \\
\cdashline{3-13}
&& 1.95 & 1.78 & 1.56 & 1.49 & 1.70 & 1.65 & 1.74 & 1.81 & 1.56 & 1.60 & 1.83 \\
\hline
\multirow{4}{*}{AfriqueGemma-12B} 
& Instruction Alignment Score & 1.72 & 1.94 & 1.93 & 1.70 & 1.77 & 1.80 & 1.92 & 1.83 & 1.85 & 1.86 & 1.83 \\
 & Pedagogical Completeness Score & 1.29 & 1.40 & 1.36 & 1.22 & 1.30 & 1.32 & 1.35 & 1.35 & 1.34 & 1.31 & 1.32 \\
 & Linguistic Cultural Accuracy Score & 1.73 & 1.74 & 1.96 & 1.64 & 1.77 & 1.76 & 1.94 & 1.79 & 1.83 & 1.81 & 1.80 \\
 & Coherence And Naturalness Score & 1.30 & 1.47 & 1.38 & 1.29 & 1.31 & 1.33 & 1.40 & 1.34 & 1.37 & 1.40 & 1.36 \\
\cdashline{3-13}
&& 1.51 & 1.64 & 1.66 & 1.46 & 1.54 & 1.55 & 1.65 & 1.58 & 1.60 & 1.59 & 1.60 \\
\hline
\multirow{4}{*}{AfriqueQwen-14B} 
& Instruction Alignment Score & 3.03 & 2.87 & 3.20 & 2.86 & 2.78 & 2.86 & 2.62 & 3.08 & 2.87 & 2.70 & 2.88 \\
 & Pedagogical Completeness Score & 1.89 & 1.78 & 1.95 & 1.68 & 1.78 & 1.80 & 1.73 & 1.90 & 1.83 & 1.73 & 1.81 \\
 & Linguistic Cultural Accuracy Score & 2.15 & 1.85 & 2.16 & 1.75 & 1.98 & 2.00 & 1.84 & 2.16 & 1.92 & 1.75 & 1.95 \\
 & Coherence And Naturalness Score & 2.20 & 2.35 & 2.63 & 2.24 & 2.27 & 2.33 & 2.21 & 2.14 & 2.40 & 2.23 & 2.30 \\
\cdashline{3-13}
&& 2.57 & 2.46 & 2.49 & 2.38 & 2.45 & 2.50 & 2.35 & 2.57 & 2.50 & 2.43 & 2.49 \\

\bottomrule
\end{tabular}}

\label{tab:benckmark-ratings}
\end{table*}


\section{Fine-tuning Hyperparameter Details} \label{app:param}
We fine-tune the LLMs with different fine-tuning parameters, the default ($\beta$ 0.1, batch size 1, epoch 3) and the cthe default ($\beta$ 0.1, batch size 1, epoch 3) and the custom ($\beta = 0.5$, batch size 4, epoch 10) from LlamaFactory\footnote{\url{https://github.com/hiyouga/LlamaFactory}} LLM fine-tuning framework. We also make a full fine-tuning and fine-tuning with LORA settings. $\beta$ is a standard beta parameter used for DPO \cite{rafailov2023direct}. We studied the impact of training parameters such as beta, batch size, and the number of epochs in our work. 

\noindent\textbf{Fine-tuning Parameter Effects.} Higher parameter values consistently outperforms the default. Low-resource alignment is sensitive to fine-tuning hyperparameters. The superiority of higher parameters suggests that more intensive training is necessary to overcome the high loss associated with unfamiliar language data. A lower parameter, such as $\beta$ (0.1), likely causes the model to diverge too quickly or fail to learn the preference signal effectively. The detail parameter ablations and results are in the Appendix \ref{app:addional-results}. However, the parameter with full fine-tuning is computationally very expensive.

\section{Additional Results across Dialog Types}
Table \ref{tab:dialog-type} shows results across dialog types.
\begin{table*}[h!]
\centering
\caption{Resulst across Dialog types from our best model:  Llama-3-8B + SFT + DPO and  Gemma3-12B + SFT + DPO.}

\resizebox{\textwidth}{!}{
\begin{tabular}{lcccccccccc|c}
\toprule
\textbf{Dialog type} & \textbf{amh} &\textbf{hau}& \textbf{ibo}& \textbf{lin}& \textbf{orm}& \textbf{som}& \textbf{swa}& \textbf{tir}& \textbf{yor}& \textbf{zul}& \textbf{Avg.}\\
\midrule
\multicolumn{12}{l}{\textit{Llama-3-8B + SFT + DPO Results}} \\ 

Cultural Note Integration&	67.19&	69.22&	68.55&	69.27&	69.61&	67.65&	66.63&	70.21&	66.9&	68.83&	68.48\\
Direct Q\&A&	68.75&	71.57&	69.58&	73.00&	73.44&	69.9&	68.84&	67.6&	69.62&	70.83&	70.27\\
Error Correction / Hinting&	70.31&	71.45&	74.29&	76.09&	73.58&	70.83&	71.31&	71.57&	71.32&	70.74&	72.16\\
Fill-in-the-Blank&	62.21&	67.57&	67.37&	68.02&	68.45&	66.44&	63.96&	65.76&	65.44&	67.36&	66.17\\
Grammar Explanation&	65.45&	69.24&	67.52&	71.93&	70.56&	68.13&	69.64&	66.15&	66.07&	71.11&	68.67\\
Role-play / Contextual Use&	67.46&	67.80&	71.09&	70.31&	71.58&	67.13&	69.16&	70.56&	68.30&	69.39&	69.34\\
Sentence Building&	68.87&	72.5&	73.46&	72.8&	71.94&	72.55&	72.48&	67.61&	70.77&	70.35&	71.38\\
Translation Practice&	63.98&	66.83&	71.94&	70.09&	68.75&	66.56&	65.33&	67.16&	66.83&	62.87&	66.87\\
Quiz (Multiple Choice)& 63.02&	66.69&	65.14&	68.30&	69.56&	63.31&	65.91&	64.55&	62.80&	66.27&	65.61\\
\cdashline{1-12}
\multicolumn{12}{l}{\textit{Irrelevant/wrong question inputs}} \\
\cdashline{1-12}
Irrelevant / Mixed Context&	70.40&	69.25&	71.57&	72.19&	71.84&	69.77&	68.81&	68.24&	68.53&	68.64&	69.93\\
Misspelled / Typo&	67.42&	70.25&	69.96&	71.7&	71.91&	69.9&	68.88&	69.78&	68.1&	69.29&	69.67\\
Factually Wrong Premise&	71.91&	74.44&	74.38&	76.55&	74.21&	71.31&	73.11&	72.39&	73.5&	71.88&	73.39\\
Out-of-Scope / Nonsensical&	72.99&	71.15&	71.32&	72.56&	71.17&	71.43&	72.46&	73.49&	70.01&	72.39&	71.98\\
Vague / Ambiguous&	68.39&	70.7&	72.24&	75.08&	73.07&	71.73&	71.45&	69.19&	68.86&	71.16&	71.1\\
\midrule
\textbf{Average}&	68.58&	70.14&	70.88&	72.41&	71.66&	69.41&	69.46&	69.48&	68.79&	69.61& 70.03\\
\midrule
\multicolumn{12}{l}{\textit{Gemma3-12B + SFT + DPO Results}} \\
Cultural Note Integration&	61.59&	62.97&	62.09&	64.45&	65.75&	63.27&	61.83&	63.83&	59.52&	61.62&	62.78\\
Direct Q\&A&	61.72&	66.54&	63.02&	62.75&	66.50&	67.22&	68.56&	62.12&	58.65&	64.09&	64.09\\
Error Correction / Hinting&	68.39&	70.88&	69.74&	72.83&	72.05&	71.40&	70.08&	66.79&	69.53&	69.13&	70.17\\
Fill-in-the-Blank&	62.02&	63.68&	67.80&	67.29&	66.77&	66.17&	67.29&	63.80&	66.27&	65.69&	65.72\\
Grammar Explanation&	53.98&	54.90&	57.48&	58.33&	58.33&	58.45&	55.02&	55.38&	56.10&	55.31&	56.43\\
Role-play / Contextual Use&	61.51&	66.98&	66.41&	67.68&	66.41&	65.97&	70.90&	62.22&	63.84&	67.35&	65.99\\
Sentence Building&	62.38&	65.52&	66.56&	64.47&	67.69&	66.03&	70.16&	65.34&	63.91&	66.35&	65.90\\
Translation Practice&	57.24&	58.17&	60.20&	59.15&	61.66&	59.87&	65.15&	60.80&	57.98&	56.90&	59.81\\
Quiz (Multiple Choice)&	65.89&	67.11&	66.53&	63.24&	67.59&	66.55&	66.82&	66.52&	64.88&	60.73&	65.74\\
\cdashline{1-12}
\multicolumn{12}{l}{\textit{Irrelevant/wrong question inputs}} \\
\cdashline{1-12}
Irrelevant / Mixed Context&	65.93&	67.19&	65.56&	67.26&	68.61&	63.52&	67.46&	63.07&	62.86&	66.46&	65.82\\
Misspelled / Typo&	61.75&	66.19&	64.14&	66.35&	69.14&	66.50&	66.76&	63.34&	64.68&	64.70&	65.30\\
Factually Wrong Premise&	68.94&	73.62&	70.68&	73.21&	75.67&	73.69&	75.91&	67.37&	70.73&	71.52&	72.01\\
Out-of-Scope / Nonsensical&	65.41&	71.51&	69.79&	72.72&	71.52&	72.39&	73.94&	67.08&	65.38&	71.30&	70.13\\
Vague / Ambiguous&	60.76&	67.32&	65.07&	63.47&	67.00&	68.15&	67.46&	61.81&	64.57&	63.78&	65.01\\
\midrule
\textbf{Average}&	63.19&	66.64&	65.66&	66.92&	68.29&	66.79&	68.24&	63.87&	64.08&	65.23&	65.90\\
\bottomrule
\end{tabular}
}

\label{tab:dialog-type}
\end{table*}

\section{Additional Automatic Evaluation Metric Results}\label{app:auto-metrics}
Table \ref{tab:auto-metrics} shows results from automatic evaluation metrics (BERTScore, ChrF++, and ROUGE-L).
\begin{table*}[h!]
\centering
\caption{Evaluation results using automatic evaluation metrics across the baseline LLMs and various fine-tuning settings such as DPO, SFT, and SFT + DPO. The DPO rejected is generated from Gemma-2-2B. }

\resizebox{\textwidth}{!}{
\begin{tabular}{llcccccccccc|c}
\toprule
\textbf{Model} &\textbf{Metrics}& \textbf{amh} &\textbf{hau}& \textbf{ibo}& \textbf{lin}& \textbf{orm}& \textbf{som}& \textbf{swa}& \textbf{tir}& \textbf{yor}& \textbf{zul}& \textbf{Avg}\\
\midrule
\multirow{3}{*}{{Llama-3.2-3B}} 
&BERTScore F1 & 0.85 & 0.85 & 0.85 & 0.85 & 0.85 & 0.85 & 0.85 & 0.85 & 0.85 & 0.85 & 0.85\\
&ChrF++ & 25.17 & 25.87 & 25.96 & 26.35 & 25.33 & 25.05 & 25.97 & 24.56 & 25.48 & 25.51 &25.52\\
&ROUGE-L & 0.18 & 0.17 & 0.17 & 0.18 & 0.17 & 0.17 & 0.18 & 0.18 & 0.17 & 0.17 & 0.17 \\
\midrule
\multirow{3}{*}{{Llama-3.2-3B + DPO}} 
&BERTScore F1 & 0.86 & 0.86 & 0.86 & 0.86 & 0.86 & 0.86 & 0.86 & 0.86 & 0.86 & 0.86 & 0.86 \\
&ChrF++ & 28.37 & 29.88 & 30.20 & 30.67 & 29.82 & 29.43 & 30.48 & 27.95 & 30.01 & 29.05 & 29.59 \\
&ROUGE-L & 0.20 & 0.20 & 0.20 & 0.20 & 0.20 & 0.19 & 0.20 & 0.21 & 0.20 & 0.19 & 0.20 \\
\midrule
\multirow{3}{*}{{Llama-3.2-3B + SFT}} 
&BERTScore F1 & 0.87 & 0.88 & 0.88 & 0.89 & 0.88 & 0.88 & 0.88 & 0.88 & 0.87 & 0.88 & 0.88 \\
&ChrF++ & 28.03 & 27.98 & 28.25 & 30.98 & 27.98 & 28.54 & 29.05 & 28.17 & 27.66 & 28.51 & 28.48 \\
&ROUGE-L & 0.26 & 0.26 & 0.26 & 0.29 & 0.26 & 0.27 & 0.27 & 0.27 & 0.26 & 0.26 & 0.27 \\
\midrule
\multirow{3}{*}{{Llama-3.2-3B +SFT+DPO}} 
&BERTScore F1 & 0.88 & 0.88 & 0.88 & 0.89 & 0.88 & 0.88 & 0.88 & 0.88 & 0.87 & 0.88 & 0.88 \\
&ChrF++ & 30.23 & 30.21 & 30.87 & 33.28 & 30.45 & 30.78 & 30.72 & 30.84 & 30.19 & 30.55 & 30.79 \\
&ROUGE-L & 0.26 & 0.26 & 0.26 & 0.29 & 0.26 & 0.27 & 0.27 & 0.27 & 0.26 & 0.26 & 0.27 \\
\midrule
\multirow{3}{*}{{Gemma3-12B}} 
&BERTScore F1 & 0.84 & 0.84 & 0.84 & 0.84 & 0.84 & 0.84 & 0.84 & 0.84 & 0.84 & 0.84 & 0.84 \\
&ChrF++ & 27.05 & 26.27 & 27.76 & 26.68 & 26.65 & 26.45 & 26.86 & 26.93 & 27.19 & 26.27 & 26.82 \\
&ROUGE-L & 0.17 & 0.16 & 0.17 & 0.17 & 0.16 & 0.16 & 0.17 & 0.17 & 0.17 & 0.16 & 0.17 \\
\midrule
\multirow{3}{*}{{Gemma3-12B+DPO}} 
&BERTScore F1 & 0.84 & 0.84 & 0.84 & 0.85 & 0.84 & 0.84 & 0.85 & 0.85 & 0.84 & 0.84 & 0.84 \\
&ChrF++ & 28.65 & 27.50 & 29.05 & 28.06 & 27.88 & 27.60 & 28.15 & 28.52 & 28.38 & 27.49 & 28.14 \\
&ROUGE-L & 0.17 & 0.16 & 0.17 & 0.17 & 0.16 & 0.16 & 0.17 & 0.17 & 0.17 & 0.16 & 0.17 \\
\midrule
\multirow{3}{*}{{Gemma3-12B+SFT}} 
&BERTScore F1 & 0.87 & 0.87 & 0.87 & 0.88 & 0.88 & 0.87 & 0.88 & 0.87 & 0.87 & 0.87 & 0.87 \\
&ChrF++ & 34.84 & 32.99 & 34.43 & 35.82 & 33.23 & 33.15 & 34.08 & 34.75 & 33.45 & 32.80 & 33.97 \\
&ROUGE-L & 0.24 & 0.26 & 0.25 & 0.27 & 0.26 & 0.26 & 0.26 & 0.25 & 0.25 & 0.24 & 0.26 \\
\midrule
\multirow{3}{*}{{Gemma3-12B+SFT+DPO}} 
&BERTScore F1 & 0.87 & 0.87 & 0.87 & 0.88 & 0.87 & 0.87 & 0.88 & 0.87 & 0.87 & 0.87 & 0.87 \\
&ChrF++ & 36.08 & 35.02 & 36.62 & 37.55 & 35.42 & 35.87 & 36.06 & 36.01 & 35.67 & 34.71 & 35.90 \\
&ROUGE-L & 0.24 & 0.25 & 0.25 & 0.26 & 0.25 & 0.25 & 0.25 & 0.24 & 0.25 & 0.23 & 0.25 \\

\bottomrule
\end{tabular}
}
\label{tab:auto-metrics}
\end{table*}
\end{document}